\documentclass{ecai}  

\usepackage{graphicx}
\usepackage{latexsym}

\usepackage{times}
\usepackage{soul}
\usepackage{url}
\usepackage[hidelinks]{hyperref}
\usepackage[utf8]{inputenc}
\usepackage[small]{caption}
\usepackage{graphicx}
\usepackage{amsmath}
\usepackage{amsthm}
\usepackage{booktabs}
\usepackage{algorithm}
\usepackage{algorithmic}
\usepackage[switch]{lineno}

\usepackage{color}
\usepackage{hyperref}
\hypersetup{
    colorlinks=false,
    linkcolor=red,
    filecolor=magenta,      
    urlcolor=magenta,
    citecolor=green,
}
\usepackage{multirow} 
\usepackage{amssymb}
\usepackage{bbding}
\usepackage{bbm}
\usepackage{stfloats}
\usepackage{cite} 

\begin{document}

\begin{frontmatter}

\title{MonoSKD: General Distillation Framework for Monocular 3D Object Detection via Spearman Correlation Coefficient}

\author[A]{\fnms{Sen}~\snm{Wang}}
\author[A]{\fnms{Jin}~\snm{Zheng}
    \thanks{Corresponding Author. Email: JinZheng@buaa.edu.cn}
} 

\address[A]{School of Computer Science and Engineering, Beihang University, Beijing, China, 100191}


\begin{abstract}
    Monocular 3D object detection is an inherently ill-posed problem, as it is challenging to predict accurate 3D localization from a single image.
    Existing monocular 3D detection knowledge distillation methods usually project the LiDAR onto the image plane and train the teacher network accordingly.
    Transferring LiDAR-based model knowledge to RGB-based models is more complex, so a general distillation strategy is needed.
    To alleviate cross-modal problem, we propose \textbf{MonoSKD}, a novel \textbf{K}nowledge \textbf{D}istillation framework for \textbf{Mono}cular 3D detection based on \textbf{S}pearman correlation coefficient, to learn the relative correlation between cross-modal features.
    Considering the large gap between these features, strict alignment of features may mislead the training, so we propose a looser Spearman loss.
    Furthermore, by selecting appropriate distillation locations and removing redundant modules, our scheme saves more GPU resources and trains faster than existing methods.
    Extensive experiments are performed to verify the effectiveness of our framework on the challenging KITTI 3D object detection benchmark.
    Our method achieves state-of-the-art performance until submission with no additional inference computational cost.
    Our codes are available at \href{https://github.com/Senwang98/MonoSKD}{https://github.com/Senwang98/MonoSKD}.
\end{abstract}

\end{frontmatter}

\section{Introduction}
Due to its widespread applications, 3D object detection has attracted significant attention in augmented reality, autonomous driving, and robot navigation.
Accurate 3D localization is the basis for ensuring security, so the key to 3D object detection is to obtain accurate 3D localization.
According to the input resources, the existing 3D object detectors can be divided into four categories: 
LiDAR point cloud-based~\cite{PointPillars}, stereo image-based~\cite{LiGA-Stereo}, 
monocular image-based~\cite{MonoFlex,CaDDN} and multi-modality-based methods~\cite{MVX-Net}.
The industry usually chooses LiDAR point cloud-based, stereo image-based, or multi-modality-based methods because they can directly perceive the depth information of surroundings.
Compared with LiDAR sensors and stereo cameras, monocular cameras are low-cost and flexible for deployment.
Considering the above unique advantages, the community has begun to pay more attention to monocular 3D object detection.

\begin{figure}[t]
    \centering {
        \includegraphics[width=0.45\textwidth]{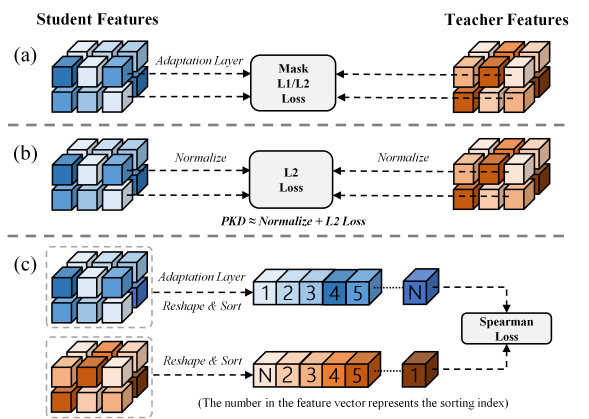}
    }
    \caption{
        \textbf{Comparison of different distillation strategies.}
        (a)~The loss of mask L1 used in MonoDistill~\cite{Monodistill}, where the mask is used to filter the background.
        (b)~The Pearson loss proposed in PKD~\cite{PKD} is equivalent to normalization combined with L2 loss.
        (c)~Our Spearman distillation loss for monocular 3D detection.
    }
    \label{fig1}
\end{figure}

Notable progress in monocular 3D object detection has been achieved in recent years.
Nevertheless, there still exists a considerable performance gap between the monocular image-based and LiDAR-based methods.
Compared with the direct acquisition of accurate depth by LiDAR, predicting depth from monocular images is an inherently ill-posed problem~\cite{PatchNet}.
To mitigate this issue, several works~\cite{PatchNet} take monocular depth estimation networks to provide dense depth maps.
Recently, some works~\cite{MonoDETR,MonoDTR} view depth estimation as an auxiliary task to introduce depth-aware features for object detectors, achieving remarkable performance improvement.
However, these detectors are still not robust enough for depth estimation, resulting in inevitable depth estimation errors.

\begin{figure*}[t]
    \centering {
        \includegraphics[width=\textwidth]{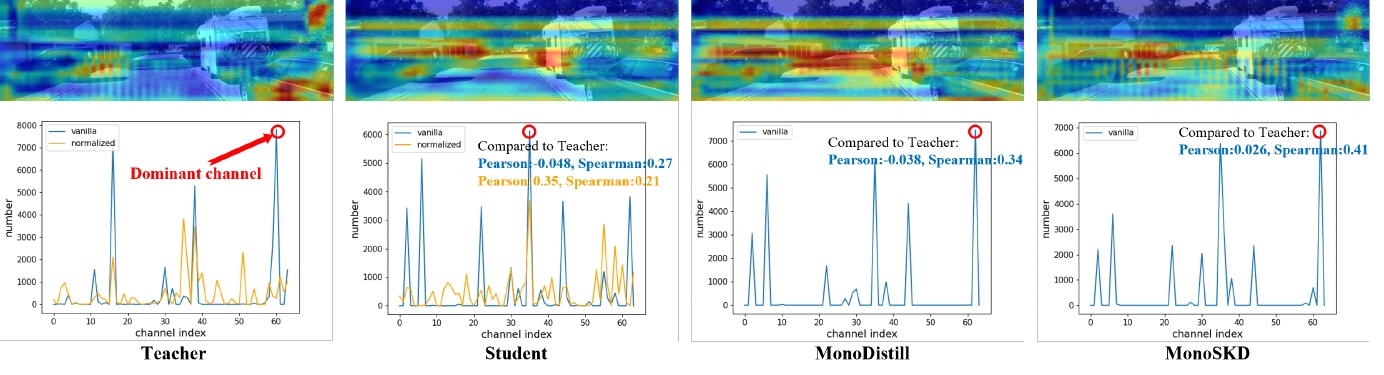}
    }
    \caption{
        \textbf{Visualization of the feature maps and dominant channels.}
        \textbf{Top}:~Visualization of teacher and student's neck feature maps.
        \textbf{Bottom}:~Dominant channels in neck stage ’P2’.
        Let $s_{l,u,v} \in  \mathbb{R}^{C}$ denote the feature vector of pixel $(u, v)$ from $l$-th neck stage and omit $l$ for clarity.
        Then $number_i=$ 
        $ \sum_{u,v} {\mathbbm{1}[{\rm argmax}_{c} \, s_{u,v}^{(c)}=i]} $
        where $i$ denotes the channel index.
}
    \label{Visualization of the multi-scale feature maps}
\end{figure*}

To alleviate the above problem, MonoDistill~\cite{Monodistill} proposes a monocular 3D detection knowledge distillation (KD) framework to transfer the depth-related knowledge of LiDAR signals to the RGB student model to improve the robustness of the detector.
MonoDistill projects the LiDAR signals into the image plane to generate the depth maps and trains the teacher model.
In this case, directly forcing the teacher and student to align at pixel level on the feature maps is suboptimal because cross-modal differences will mislead training.
Here, LiDAR signals are only used for training, and the student network is a standard monocular 3D detection network during the inference stage.
Moreover, PKD~\cite{PKD} has proven that applying normalization in 2D object detection can bridge the feature gap between the student and the teacher.
However, due to vast modal differences, PKD only has a slight improvement in the cross-modal task.
In Figure \ref{fig1}~(a) and (b), we show the two feature distillation strategies adopted by MonoDistill and PKD.
The Mask L1 loss used in MonoDistill directly aligns the foreground regions of the feature maps from teacher and student, and PKD aligns normalized feature maps.
At the top of Figure \ref{Visualization of the multi-scale feature maps}, the visualized feature maps show huge differences between teacher and student, especially in foreground regions. 
In contrast, the visual feature map of MonoSKD is closer to that of the teacher.
The blue curve at the bottom of Figure \ref{Visualization of the multi-scale feature maps} represents the dominant channel of the naive feature maps.
The teacher's and student's dominant channels show the vast channel difference.
Meanwhile, we migrate the PKD distillation strategy to the monocular 3D detection field and use the yellow curve to represent the dominant channel of the normalized feature maps.
Compared to the student, MonoDistill and MonoSKD are more similar to teachers on the blue curve.
For quantitative evaluation, we choose Pearson and Spearman correlation coefficients as quantitative metrics and compare student, MonoDistill, and MonoSKD with the teacher's dominant channel curve. 
The closer the metrics are to 1, the better.
Even though the feature difference after normalization is reduced (-0.048 $\rightarrow$ 0.35), it still faces vast channel differences.
Normalization will destroy the dominant channel ranking relationship, so the Spearman correlation coefficient drops (0.27 $\rightarrow$ 0.21).
Therefore, although PKD has improved on the Pearson metric, it can disrupt the ranking relationship between features, thereby reducing the Spearman metric.
Compared with MonoDistill, the MonoSKD proposed in this paper is more similar to the teacher's dominant channel.
MonoSKD achieves better results than MonoDistill on both metrics because feature maps of different modalities have significant pixel-level differences, making it challenging to satisfy strict pixel alignment.

Considering the above deficiencies, we try to mine more general knowledge of cross-modal features, such as relative relationships.
Therefore, we introduce the Spearman correlation coefficient~(SCC) in the monocular 3D detection distillation framework to learn the ranking relationship between cross-modal features.
In Figure \ref{fig1}~(c), we show the process of Spearman loss. 
It can be seen that Spearman loss only cares about sorting information between features rather than specific values, which is more suitable for cross-modal tasks.

Besides, we find that the existing distillation framework suffers from redundant distillation modules. 
We select the appropriate distillation location and remove redundant modules, which saves about 30\% of the average GPU memory usage, accelerates training, and improves distillation performance.

To verify our scheme's effectiveness and generality, we perform distillation experiments on three recent monocular 3D detectors, including MonoDLE~\cite{MonoDLE}, GUPNet~\cite{GUPNet}, and DID-M3D~\cite{DID-M3D}.
As expected, our method dramatically improves the performance of these three detectors on the KITTI~\cite{Kitti} benchmark.

In summary, our contributions are listed as follows:
 
\begin{itemize}

\item[$\bullet$] 
We propose a general Spearman distillation strategy for the knowledge distillation task of monocular 3D detection to learn the ranking relationship between features and improve performance.
\item[$\bullet$]
We find that MonoDistill suffers from redundant distillation modules, and our redesigned distillation framework saves an average of 30\% of GPU memory and accelerates training while improving distillation performance.
\item[$\bullet$]
We conduct extensive experiments on three detectors using the challenging KITTI benchmark to demonstrate the effectiveness and generality of our framework.
Our method achieves state-of-the-art performance with no extra inference computational cost.
 
\end{itemize}

\section{Related Work}
\subsection{Monocular 3D object detection}
Given an input image, monocular 3D object detection aims to predict a 3D bounding box represented by its location, dimension, and orientation for each object.
Based on whether to use additional data, existing methods can be divided into two categories: standard monocular 3D detectors and detectors using additional data.
Standard monocular 3D detection methods such as MonoDLE only use the RGB images, annotations, and camera calibrations provided by KITTI dataset~\cite{Kitti} to predict 3D bounding boxes.
Mousavian et al.~\cite{multibin} combined estimated 3D object orientation and dimensions with the geometric constraints on translation imposed by the 2D bounding box to recover 3D locations.
MonoPair~\cite{MonoPair} encoded spatial constraints for partially-occluded objects from their adjacent neighbors to improve the monocular 3D
object detection.
Qin et al. proposed MonoGRNet~\cite{MonoGRNet} for monocular 3D object detection via geometric reasoning in both the observed 2D projection and the unobserved depth dimension.
OFTNet~\cite{OFTNet} introduced an orthographic feature transform to map image-based features into an orthographic 3D space.
Instead of directly regressing depth, some works such as GUPNet predicted 2D and 3D heights via uncertainty modeling and recovered 3D locations based on geometric priors.
Meanwhile, several works~\cite{Monocon,MonoFlex} utilized keypoint-based geometric constraints to improve the monocular 3D object detection further.
These standard monocular 3D object detectors had achieved prominent progress but still suffered from the unsatisfactory performance of the 3D locations.

On the other hand, some methods use additional data.
Deep Manta~\cite{Deepmanta} improved performance by using more detailed annotated locations of key points, e.g., wheels, as training labels.
several works~\cite{ROI10d} use the CAD models as shape templates to get better object geometry. 
Specifically, AutoShape~\cite{AutoShape} generated shape-aware key points via CAD models to boost the detection performance.
Several works~\cite{PatchNet,D4LCN} accomplished monocular 3D object detectors by directly taking depth maps from off-the-shelf depth estimators as extra inputs.
Other works utilize additional data to help with online depth estimation.
For example, DD3D~\cite{DD3D} utilized a large private dataset and the KITTI depth dataset for depth pre-training to improve detection performance.
MonoDTR~\cite{MonoDTR} proposed an end-to-end transformer for monocular 3D object detection, which utilized depth maps as auxiliary supervision.
Recently, DID-M3D introduced dense depth maps to decouple the instance depth.
However, these detectors are still not robust enough due to unavoidable depth errors.

\subsection{Knowledge distillation}
The concept of knowledge distillation (KD)~\cite{KD} was first proposed for model compression, which trains student models with GT labels and soft labels from teacher networks.
Instead of transferring knowledge from teachers' responses, Romero et al.~\cite{fitnets} proved that intermediate features distillation can also guide the training of student networks.
After that, more and more tasks utilized knowledge distillation to achieve a remarkable performance improvement, such as object detection~\cite{PKD} and semantic segmentation~\cite{CWD}, etc.
Specifically, KD in object detection is usually divided into three categories: feature-based, relation-based, and response-based.
Feature-based KD usually carefully designs a mask for knowledge localization and transfers knowledge after feature alignment (such as dimension and semantic alignment).
Relation-based KD considers the difference in feature relations instead of the pixel-to-pixel difference between corresponding feature maps.
Response-based KD is a commonly used and efficient distillation method, using the teacher's prediction as a soft label to supervise the student network.

For object detection, Chen et al.~\cite{chen2017} first introduced knowledge distillation to 2D object detection, which distilled the neck, regression head, and classification head of detectors.
Afterward, Wang et.al~\cite{FGFI} proposed a fine-grained feature imitation method.
Instead of distilling foreground object regions, Guo et al.~\cite{DeKD} proposed that the distillation of background regions can also be effective.
Recently, PKD introduced Pearson correlation coefficient (PCC)~\cite{pearson} and normalization strategy for homogeneous and heterogeneous detectors.
For 3D object detection, Guo et al. proposed LiGA-Stereo~\cite{LiGA-Stereo} to learn stereo-based 3D detectors under the guidance of high-level geometry-aware representations of LiDAR-based detection models.
PointDistiller~\cite{PointDistiller} designed a structured knowledge distillation framework for point clouds-based 3D detection.
MonoDistill achieved state-of-the-art performance by proposing a teacher model based on inputs of a projected LiDAR signals to guide student detectors with spatial cues for monocular 3D object detection.
%
Nevertheless, MonoDistill's strict alignment of cross-modal features is suboptimal.

\section{Methodology}

\begin{figure*}[htbp]
    \centering {
        \includegraphics[width=0.9\textwidth]{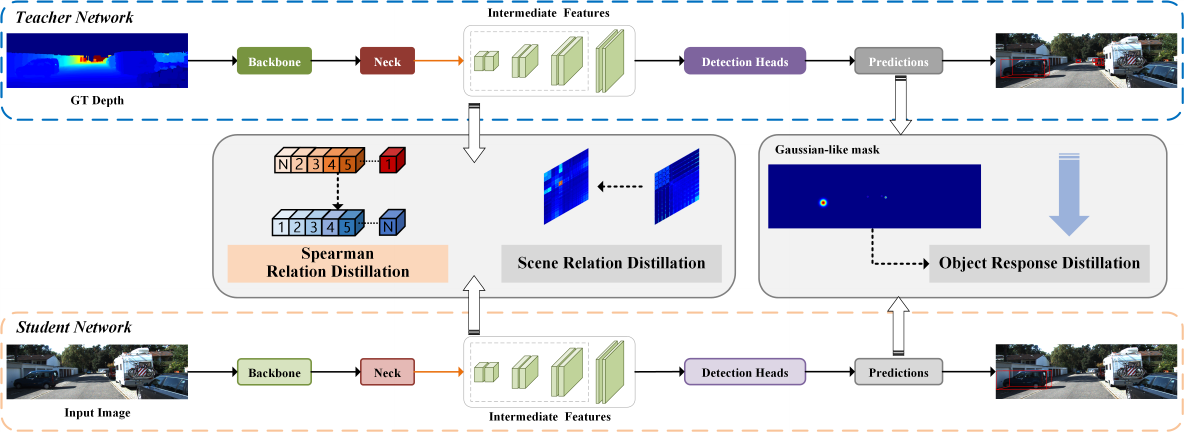}
    }
    \caption{
        \textbf{Illustration of the proposed MonoSKD.}
        The overall design follows MonoDistill.
        First, we train the teacher network offline with the processed GT depth, which shares the same architecture as the student.
        Second, we load and freeze the teacher network and guide the student network with the teacher's features and responses.
        In the inference phase, we only use the parameters of the student network.
    }
    \label{Illustration of the proposed MonoSKD}
\end{figure*}

\subsection{Overview and Framework}
General monocular 3D detection takes an image captured by an RGB camera as input, predicting 3D bounding boxes of objects for each object in 3D space.
The 3D bounding boxes are usually represented by 3D center location ($x,y,z$), dimension ($h,w,l$), and the orientation $\theta$.
Most existing monocular 3D detectors obtain 2D features through the backbone and neck and recover 3D locations through multiple independent heads.

As shown in Figure \ref{Illustration of the proposed MonoSKD}, our distillation method differs from the existing MonoDistill in the knowledge distillation loss function and the selection of distillation location.
On the one hand, we abandon the strict feature alignment distillation strategy and introduce a relation-based Spearman distillation loss, which adopts a more general distillation alignment strategy and is more suitable for cross-modal distillation. 
On the other hand, MonoDistill fuses and distills the output features of the backbone.
We find that MonoDitill has redundant distillation modules, resulting in inefficient distillation.
Thus, we use the neck output feature as the distillation object, so the heavy feature fusion module is removed, which allows us to improve performance while saving about 30\% of GPU memory and speeding up training.

\subsection{Knowledge Distillation with Spearman Correlation Coefficient}
We empirically find that the discrepancy of predictions between the RGB-based student and LiDAR-based teacher may tend to be pretty severe.
In this case, directly forcing the teacher and student to align at pixel level on the feature maps is suboptimal because cross-modal differences will mislead training.
Instead of strictly aligning features, we guide distillation training with looser constraints.
Recently, PKD introduced the Pearson correlation coefficient for object detection.
PKD has proved that applying PCC for feature maps is equivalent to normalization combined with mean square error.
The normalization mechanism bridges the gap between the activation patterns of the student and the teacher.
However, our experiment shows a small gain when applying PKD to monocular 3D detection distillation (see Appendix \ref{Discussion about PKD}).
As is shown in Figure \ref{Visualization of the multi-scale feature maps}, the activation patterns of the teacher and student networks are seriously different.
Although normalization alleviates the differences between teachers and students, the differences in the cross-modal task are still considerable.

Considering the enormous cross-modal differences, the direct alignment of specific values between feature maps is too rigorous for network training. 
To seek a looser distillation strategy, we consider the relative correlation between distillation features, so the Spearman correlation coefficient is introduced.
Such a distillation strategy can reduce the difficulty of distillation and further improve performance.
We adopt Spearman's distance as the metric, $i.e.$,
\begin{align}
    \mathcal{L}_{scc}(s,t) = \frac{1}{L} \sum_{l=1}^{L}(1 - r_{scc}(s_l,t_l))
    \label{eq1}
\end{align}
$r_{scc}$, $L$, $s$, $t$ represent the Spearman correlation coefficient, number of the feature maps from neck, student feature maps, and teacher feature maps respectively,
\begin{equation}
\label{eq2}
\begin{aligned}
r_{scc}(s, t) & = \frac{ {\rm Cov}({\rm R}(s),({\rm R}(t))}{ {\rm Std}(({\rm R}(s)) {\rm Std}(({\rm R}(t))} \\
              & = \frac{ \sum_{i=1}^{B}( {\rm R}(s_i)-{\rm \overline{R}}(s) ) ( {\rm R}(t_i)-{\rm \overline{R}}(t) ) } 
                      {
                        \sqrt{ \sum_{i=1}^{B}({\rm R}(s_i) - {\rm \overline{R}}(s))^2 }
                        \sqrt{ \sum_{i=1}^{B}({\rm R}(t_i) - {\rm \overline{R}}(t))^2 }
                      }              
\end{aligned}
\end{equation}
where 
$B$ denotes batch size, 
${\rm R}(s_i)$ is rank index of $s_i$, 
{\rm Cov}({\rm R}(s)) is the covariance of ${\rm R}(s)$, 
${\rm \overline{R}}(s)$ and ${\rm Std}(({\rm R}(s))$ 
denote the mean and standard derivation of ${\rm R}(s)$, respectively.

However, to evaluate the ranking relationship between different features, the Spearman correlation coefficient requires a ranking operation, which is not differentiable.
Fortunately, Blondel et al.~\cite{torchsort} introduced a novel method for fast differentiable sorting and ranking, so we adopted it. 

\subsection{Selection of Distillation Objects} \label{Selection of Distillation Objects}
\begin{figure}[htbp]
    \centering {
        \includegraphics[width=8.5cm]{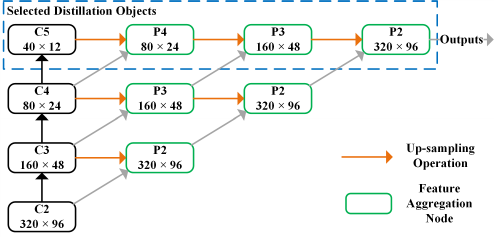}
    }
    \caption{
        \textbf{Illustration of our distillation object selection strategy.}
        The green rectangle and orange arrow denote the feature aggregation node and up-sampling operation.
        The feature maps in the blue rectangle will be distilled.
    }
    \label{dlaup}
\end{figure}

Although MonoDistill proposes a practical monocular 3D detection knowledge distillation framework, it still faces the problem of excessive GPU memory usage. Specifically, in order to more effectively distill the output features of the backbone, MonoDistill introduces an attention-based feature fusion module. We argue that this strategy is inefficient and redundant.
In contrast, we directly select the multi-scale feature maps of the neck as the distillation objects to deal with rich spatial information because the neck has the function of feature fusion.
Another empirical explanation for choosing the feature map output from the neck as the object of distillation is that shallow features are more susceptible to noise than high-level features.
Removing the heavy fusion module can save 30\% GPU memory and speed up training.
In particular, to introduce more information to aid in distillation, we keep the first layer output of the neck for distillation, which MonoDistill discards.
In our experiments, selecting the multi-scale features of the neck for distillation can also improve performance (see Table \ref{table_kitti_val_ab} for more details).
In Figure \ref{dlaup}, we show the feature fusion process of the neck and the selected objects for distillation.



\subsection{Other Knowledge Distillation Strategy}
By proposing a distillation strategy using the Spearman correlation coefficient to mine the ranking knowledge of networks across different modalities, 
we alleviate the misleading training problem faced by strict feature alignment distillation.
However, the knowledge learned by the student model still needs to be improved to support the performance requirements of the 3D detector because the ranking relationship cannot fully represent the similarity between features.
In order to obtain high-performance detectors, we still need a strict relational distillation to achieve high feature similarity in advance. 
We find that strict relation distillation combined with loose Spearman distillation can transfer the dark knowledge of the teacher model more effectively.
Based on the above considerations, we retain the scene relation distillation in MonoDistill and let it play the role of strict relational distillation.
Define $f_i$ and $f_j$ denote the $i^{th}$ and $j^{th}$ feature vector, and we calculate the similarity map as follows:
\begin{equation}
\label{eq3}
\begin{aligned}
    S_{i,j} = \frac{f_i^{\rm T} f_j}{\Vert f_i \Vert_2 \cdot  \Vert f_j \Vert_2}
\end{aligned}
\end{equation}
After that, we utilize L1 loss to train similarity maps of the teacher and student network.
\begin{equation}
\label{eq4}
\begin{aligned}
    \mathcal{L}_{sd} = \frac{1}{K \times K} \sum_{i=1}^{K} \sum_{j=1}^{K} \Vert S_{i,j}^t - S_{i,j}^s \Vert_1
\end{aligned}
\end{equation}
Where $\mathcal{L}_{sd}$, $K$, $s$, $t$ represent the scene distillation loss, the number of feature vectors, and the student and teacher feature maps, respectively.

Furthermore, like most distillation works, we use the teacher's predictions as soft labels to guide the student network training.
The object response distillation loss can be formulated as follows:
\begin{equation}
\label{eq5}
\begin{aligned}
    \mathcal{L}_{od} = \frac{1}{N} \sum_{k=1}^{N} \Vert M_{g} (p_k^s - p_k^t) \Vert_1
\end{aligned}
\end{equation}
where $N$, $M_g$, $p_k^s$ and $p_k^t$ represent the number of detection heads, the Gaussian-like mask, and prediction of the $k^{th}$ detection head from student and teacher network.

\begin{table*}
\setlength\tabcolsep{10pt}
\centering
\caption{
\textbf{Quantitative comparisons of the Car category on the KITTI \textit{testing} set.}
The best results are listed in \textbf{\textcolor[rgb]{1,0,0}{red}} and the second in \textbf{\textcolor[rgb]{0,0,1}{blue}}.
Note that DD3D employs the large private DDAD15M dataset (containing approximately 15M frames).
}
\label{table1}
\begin{tabular}{c|c|c|c|c|c|c|c|c|c|c}
\toprule
    \multicolumn{2}{c|}{\multirow{2}{*}{Methods}}
    &\multirow{2}{*}{Venue}
    &\multirow{2}{*}{Extra Data}
    &\multirow{2}{*}{Runtime}
    &\multicolumn{3}{c|}{$AP_{3D}$ (Car test)}
    &\multicolumn{3}{c}{$AP_{BEV}$ (Car test)}\\

\multicolumn{2}{c|}{} & & &
    & \multicolumn{1}{c}{Easy} 
    & \multicolumn{1}{c}{Mod.}
    & \multicolumn{1}{c|}{Hard} 
    & \multicolumn{1}{c}{Easy} 
    & \multicolumn{1}{c}{Mod.}
    & \multicolumn{1}{c}{Hard} \\
\midrule

\multicolumn{2}{l|}{MonoDLE~\cite{MonoDLE}}
    & \multicolumn{1}{c|}{CVPR 2021} 
    & \multicolumn{1}{c|}{None}
    & \multicolumn{1}{c|}{40ms}
    & \multicolumn{1}{c}{17.23} 
    & \multicolumn{1}{c}{12.26} 
    & \multicolumn{1}{c|}{10.29}
    & \multicolumn{1}{c}{24.79} 
    & \multicolumn{1}{c}{18.89}
    & \multicolumn{1}{c}{16.00} \\
    
\multicolumn{2}{l|}{MonoEF~\cite{MonoEF}}
    & \multicolumn{1}{c|}{CVPR 2021} 
    & \multicolumn{1}{c|}{None}
    & \multicolumn{1}{c|}{30ms}
    & \multicolumn{1}{c}{21.29} 
    & \multicolumn{1}{c}{13.87} 
    & \multicolumn{1}{c|}{11.71}
    & \multicolumn{1}{c}{29.03} 
    & \multicolumn{1}{c}{19.70}
    & \multicolumn{1}{c}{17.26} \\

\multicolumn{2}{l|}{MonoFlex~\cite{MonoFlex}}
    & \multicolumn{1}{c|}{CVPR 2021} 
    & \multicolumn{1}{c|}{None}
    & \multicolumn{1}{c|}{35ms}
    & \multicolumn{1}{c}{19.94} 
    & \multicolumn{1}{c}{12.89} 
    & \multicolumn{1}{c|}{12.07}
    & \multicolumn{1}{c}{28.23} 
    & \multicolumn{1}{c}{19.75}
    & \multicolumn{1}{c}{16.89} \\

\multicolumn{2}{l|}{MonoRCNN~\cite{MonoRCNN}}
    & \multicolumn{1}{c|}{ICCV 2021} 
    & \multicolumn{1}{c|}{None}
    & \multicolumn{1}{c|}{70ms}
    & \multicolumn{1}{c}{18.36} 
    & \multicolumn{1}{c}{12.65} 
    & \multicolumn{1}{c|}{10.03}
    & \multicolumn{1}{c}{25.48} 
    & \multicolumn{1}{c}{18.11}
    & \multicolumn{1}{c}{14.10} \\

\multicolumn{2}{l|}{GUPNet~\cite{GUPNet}}
    & \multicolumn{1}{c|}{ICCV 2021} 
    & \multicolumn{1}{c|}{None}
    & \multicolumn{1}{c|}{34ms}
    & \multicolumn{1}{c}{22.26} 
    & \multicolumn{1}{c}{15.02} 
    & \multicolumn{1}{c|}{13.12}
    & \multicolumn{1}{c}{30.29} 
    & \multicolumn{1}{c}{21.19}
    & \multicolumn{1}{c}{18.20} \\

\multicolumn{2}{l|}{MonoCon~\cite{Monocon}}
    & \multicolumn{1}{c|}{AAAI 2022} 
    & \multicolumn{1}{c|}{None}
    & \multicolumn{1}{c|}{26ms}
    & \multicolumn{1}{c}{22.50} 
    & \multicolumn{1}{c}{16.46} 
    & \multicolumn{1}{c|}{13.95}
    & \multicolumn{1}{c}{31.12} 
    & \multicolumn{1}{c}{22.10}
    & \multicolumn{1}{c}{19.00} \\

\midrule
\multicolumn{2}{l|}{Kinematic3D~\cite{Kinematic3D}}
    & \multicolumn{1}{c|}{ECCV 2020} 
    & \multicolumn{1}{c|}{Temporal}
    & \multicolumn{1}{c|}{120ms}
    & \multicolumn{1}{c}{19.07} 
    & \multicolumn{1}{c}{12.72} 
    & \multicolumn{1}{c|}{9.17}
    & \multicolumn{1}{c}{26.69} 
    & \multicolumn{1}{c}{17.52}
    & \multicolumn{1}{c}{13.10} \\
    
\multicolumn{2}{l|}{AutoShape~\cite{AutoShape}}
    & \multicolumn{1}{c|}{ICCV 2021} 
    & \multicolumn{1}{c|}{CAD}
    & \multicolumn{1}{c|}{40ms}
    & \multicolumn{1}{c}{22.47} 
    & \multicolumn{1}{c}{14.17} 
    & \multicolumn{1}{c|}{11.36}
    & \multicolumn{1}{c}{30.66} 
    & \multicolumn{1}{c}{20.08}
    & \multicolumn{1}{c}{15.95} \\

\multicolumn{2}{l|}{PatchNet~\cite{PatchNet}}
    & \multicolumn{1}{c|}{ECCV 2020} 
    & \multicolumn{1}{c|}{LiDAR}
    & \multicolumn{1}{c|}{400ms}
    & \multicolumn{1}{c}{15.68} 
    & \multicolumn{1}{c}{11.12} 
    & \multicolumn{1}{c|}{10.17}
    & \multicolumn{1}{c}{22.97} 
    & \multicolumn{1}{c}{16.86}
    & \multicolumn{1}{c}{14.97} \\

\multicolumn{2}{l|}{D4LCN~\cite{D4LCN}}
    & \multicolumn{1}{c|}{CVPR 2020} 
    & \multicolumn{1}{c|}{LiDAR}
    & \multicolumn{1}{c|}{200ms}
    & \multicolumn{1}{c}{16.65} 
    & \multicolumn{1}{c}{11.72} 
    & \multicolumn{1}{c|}{9.51}
    & \multicolumn{1}{c}{22.51} 
    & \multicolumn{1}{c}{16.02}
    & \multicolumn{1}{c}{12.55} \\

\multicolumn{2}{l|}{DDMP-3D~\cite{DDMP-3D}}
    & \multicolumn{1}{c|}{CVPR 2021} 
    & \multicolumn{1}{c|}{LiDAR}
    & \multicolumn{1}{c|}{180ms}
    & \multicolumn{1}{c}{19.71} 
    & \multicolumn{1}{c}{12.78} 
    & \multicolumn{1}{c|}{9.80}
    & \multicolumn{1}{c}{28.08} 
    & \multicolumn{1}{c}{17.89}
    & \multicolumn{1}{c}{13.44} \\

\multicolumn{2}{l|}{CaDDN~\cite{CaDDN}}
    & \multicolumn{1}{c|}{CVPR 2021} 
    & \multicolumn{1}{c|}{LiDAR}
    & \multicolumn{1}{c|}{630ms}
    & \multicolumn{1}{c}{19.17} 
    & \multicolumn{1}{c}{13.41} 
    & \multicolumn{1}{c|}{11.46}
    & \multicolumn{1}{c}{27.94} 
    & \multicolumn{1}{c}{18.91}
    & \multicolumn{1}{c}{17.19} \\

\multicolumn{2}{l|}{MonoDTR~\cite{MonoDTR}}
    & \multicolumn{1}{c|}{CVPR 2022} 
    & \multicolumn{1}{c|}{LiDAR}
    & \multicolumn{1}{c|}{37ms}
    & \multicolumn{1}{c}{21.99} 
    & \multicolumn{1}{c}{15.39} 
    & \multicolumn{1}{c|}{12.73}
    & \multicolumn{1}{c}{28.59} 
    & \multicolumn{1}{c}{20.38}
    & \multicolumn{1}{c}{17.14} \\
    
\multicolumn{2}{l|}{MonoDistill~\cite{Monodistill}}
    & \multicolumn{1}{c|}{ICLR 2022} 
    & \multicolumn{1}{c|}{LiDAR}
    & \multicolumn{1}{c|}{40ms}
    & \multicolumn{1}{c}{22.97} 
    & \multicolumn{1}{c}{16.03} 
    & \multicolumn{1}{c|}{13.60}
    & \multicolumn{1}{c}{31.87} 
    & \multicolumn{1}{c}{22.59}
    & \multicolumn{1}{c}{19.72} \\

\multicolumn{2}{l|}{DID-M3D~\cite{DID-M3D}}
    & \multicolumn{1}{c|}{ECCV 2022} 
    & \multicolumn{1}{c|}{LiDAR}
    & \multicolumn{1}{c|}{40ms}
    & \multicolumn{1}{c}{24.40}
    & \multicolumn{1}{c}{16.29} 
    & \multicolumn{1}{c|}{13.75}
    & \multicolumn{1}{c}{32.95}
    & \multicolumn{1}{c}{22.76}
    & \multicolumn{1}{c}{19.83} \\


\multicolumn{2}{l|}{DD3D~\cite{DD3D}}
    & \multicolumn{1}{c|}{ICCV 2021} 
    & \multicolumn{1}{c|}{External}
     & \multicolumn{1}{c|}{-}
    & \multicolumn{1}{c}{23.22} 
    & \multicolumn{1}{c}{16.34} 
    & \multicolumn{1}{c|}{14.20}
    & \multicolumn{1}{c}{32.35} 
    & \multicolumn{1}{c}{23.41}
    & \multicolumn{1}{c}{20.42} \\

\multicolumn{2}{l|}{CMKD~\cite{CMKD}}
    & \multicolumn{1}{c|}{ECCV 2022} 
    & \multicolumn{1}{c|}{External}
    & \multicolumn{1}{c|}{630ms}
    & \multicolumn{1}{c}{\textbf{\textcolor[rgb]{0,0,1}{25.09}}}
    & \multicolumn{1}{c}{16.99} 
    & \multicolumn{1}{c|}{\textbf{\textcolor[rgb]{1,0,0}{15.30}}}
    & \multicolumn{1}{c}{33.69}
    & \multicolumn{1}{c}{23.10}
    & \multicolumn{1}{c}{\textbf{\textcolor[rgb]{1,0,0}{20.67}}} \\

\midrule
\multicolumn{2}{l|}{\multirow{1}{*}{\textbf{MonoSKD+MonoDLE}}}
    & \multicolumn{1}{c|}{-} 
    & \multicolumn{1}{c|}{LiDAR}
    & \multicolumn{1}{c|}{40ms}
    & \multicolumn{1}{c}{24.75}
    & \multicolumn{1}{c}{\textbf{\textcolor[rgb]{0,0,1}{17.07}}}
    & \multicolumn{1}{c|}{14.41}
    & \multicolumn{1}{c}{\textbf{\textcolor[rgb]{0,0,1}{34.43}}}
    & \multicolumn{1}{c}{\textbf{\textcolor[rgb]{0,0,1}{23.62}}}
    & \multicolumn{1}{c}{\textbf{\textcolor[rgb]{0,0,1}{20.59}}} \\
\multicolumn{2}{l|}{\multirow{1}{*}{Improvements (to baseline)}}
    & \multicolumn{1}{c|}{} 
    & \multicolumn{1}{c|}{}
    & \multicolumn{1}{c|}{}
    & \multicolumn{1}{c}{+7.52} 
    & \multicolumn{1}{c}{+4.81} 
    & \multicolumn{1}{c|}{+4.12}
    & \multicolumn{1}{c}{+9.64} 
    & \multicolumn{1}{c}{+4.73}
    & \multicolumn{1}{c}{+4.59} \\

\midrule
\multicolumn{2}{l|}{\multirow{1}{*}{\textbf{MonoSKD+DID-M3D}}}
    & \multicolumn{1}{c|}{-} 
    & \multicolumn{1}{c|}{LiDAR}
    & \multicolumn{1}{c|}{40ms}
    & \multicolumn{1}{c}{\textbf{\textcolor[rgb]{1,0,0}{28.43}}}
    & \multicolumn{1}{c}{\textbf{\textcolor[rgb]{1,0,0}{17.35}}}
    & \multicolumn{1}{c|}{\textbf{\textcolor[rgb]{0,0,1}{15.01}}}
    & \multicolumn{1}{c}{\textbf{\textcolor[rgb]{1,0,0}{37.12}}}
    & \multicolumn{1}{c}{\textbf{\textcolor[rgb]{1,0,0}{24.08}}}
    & \multicolumn{1}{c}{20.37}\\
\multicolumn{2}{l|}{\multirow{1}{*}{Improvements (to baseline)}}
    & \multicolumn{1}{c|}{} 
    & \multicolumn{1}{c|}{}
    & \multicolumn{1}{c|}{}
    & \multicolumn{1}{c}{+4.03} 
    & \multicolumn{1}{c}{+1.06} 
    & \multicolumn{1}{c|}{+1.26}
    & \multicolumn{1}{c}{+4.17} 
    & \multicolumn{1}{c}{+1.32}
    & \multicolumn{1}{c}{+0.54} \\

\bottomrule
\end{tabular}
\label{table_kitti_test}
\end{table*}

\subsection{End-to-end Training}
For the convenience of expression, we define the inherited losses from the student network as $\mathcal{L}_{reg}$, $\mathcal{L}_{cls}$ and $\mathcal{L}_{dep}$  for bounding box regression, object classification, and depth regression, respectively.
The total training loss is:
\begin{equation}
\begin{aligned}
    \mathcal{L} = \mathcal{L}_{reg} + \mathcal{L}_{cls} + \mathcal{L}_{dep} + \mathcal{L}_{od} + \mathcal{L}_{sd} + \alpha\mathcal{L}_{scc}
\end{aligned}
\label{eq6}
\end{equation}
where $\alpha$ are hyper-parameters to balance the detection training loss and distillation loss.
Rather than specially selecting the optimal hyper-parameters, we choose $\alpha=1$ for simplicity.

\section{Experiments}
\subsection{Dataset and Metrics}
Following the previous works~\cite{MonoDLE,GUPNet}, we perform experiments on the challenging KITTI~\cite{Kitti} dataset.
The KITTI dataset comprises 7,481 training samples and 7,518 testing samples, where the labels of training samples are publicly available, and the labels of testing samples are private, which are only used for online evaluation and ranking.
To conduct ablations, we further divide the training samples into a train set (3,712 samples) and a validation set (3,769 samples), following prior works~\cite{3DOP}.
The final results of our method are reported on \textit{testing} set, while the ablation studies are conducted on the \textit{validation} set.
Besides, KITTI divides objects into \textit{easy}, \textit{moderate}, and \textit{hard} levels according to the 2D box height, occlusion, and truncation levels of one object.
In KITTI, only three categories of performance are mainly concerned: car, pedestrian, and cyclist, among which the performance of \textbf{car} with the \textbf{moderate} level is the most critical.
For evaluation metrics, both 3D detection and Bird’s Eye View (BEV) detection are evaluated using the $AP_{40}$ metric.


\subsection{Implementation Details}
To demonstrate the generalizability and effectiveness of our method, we choose three monocular 3D detection networks for distillation experiments: MonoDLE, GUPNet, and DID-M3D.
We accomplish our method with the PyTorch framework~\cite{pytorch}.
Taking the MonoDLE detector as an example, we conduct experiments on 2 NVIDIA RTX 3090 GPUs with batch size 12 and train it for 160 epochs, which takes almost 10 hours.
See Appendix \ref{More Details of Our Experiments} for experimental details of GUPNet and DID-M3D.
We choose the Adam optimizer with the initial learning rate $1e^{-5}$.
We apply the linear warm-up strategy for the first five epochs of training, which increases the learning rate to $1e^{-3}$.
Afterward, the learning rate decays in epochs 95 and 125 with a rate of 0.1.
For the backbone, neck, and head, we follow the design of MonoDistill.
We train the teachers with the same dense depth maps used in MonoDistill for a fair comparison.
To execute the distillation process, we have pre-trained the teacher network, and the teacher performance is demonstrated in Appendix Table \ref{table_stu_tea}.
Additionally, our code will be open-sourced for reproducibility.

\begin{table}
\setlength\tabcolsep{4.4pt}
\centering
\caption{
\textbf{Ablation studies on the KITTI validation set.}
We conduct experiments based on the \textbf{MonoDLE} network.
SD, SCC, and OD denote the scene distillation, the Spearman distillation, and the object response distillation, respectively.
$\dagger$:~Distilled object selection strategy proposed in Section \ref{Selection of Distillation Objects}.
}
\label{table1}
\begin{tabular}{c|ccc|c|c|c|c|c|c}
\toprule
    \multicolumn{1}{c|}{\multirow{2}{*}{\#}}
    &\multirow{2}{*}{SD}
    &\multirow{2}{*}{SCC}
    &\multirow{2}{*}{OD}
    &\multicolumn{3}{c|}{$AP_{3D}$ (Car val)}
    &\multicolumn{3}{c}{$AP_{BEV}$ (Car val)}\\

\multicolumn{1}{c|}{}
&
&
&
& \multicolumn{1}{c}{Easy} 
& \multicolumn{1}{c}{Mod.} 
& \multicolumn{1}{c|}{Hard} 
& \multicolumn{1}{c}{Easy} 
& \multicolumn{1}{c}{Mod.}
& \multicolumn{1}{c}{Hard} \\
\midrule

\multicolumn{4}{c|}{MonoDistill}
    & \multicolumn{1}{c}{24.40} 
    & \multicolumn{1}{c}{18.47} 
    & \multicolumn{1}{c|}{16.46}
    & \multicolumn{1}{c}{32.86} 
    & \multicolumn{1}{c}{25.14}
    & \multicolumn{1}{c}{21.99}\\

\midrule
\multicolumn{4}{c|}{Our MonoDistill$^\dagger$}
    & \multicolumn{1}{c}{24.67} 
    & \multicolumn{1}{c}{18.64} 
    & \multicolumn{1}{c|}{15.74}
    & \multicolumn{1}{c}{34.40} 
    & \multicolumn{1}{c}{25.81} 
    & \multicolumn{1}{c}{22.45}\\ 

\midrule
\multicolumn{1}{l|}{(a)}
&
&
&
& \multicolumn{1}{c}{19.86} 
& \multicolumn{1}{c}{15.11} 
& \multicolumn{1}{c|}{12.64}
& \multicolumn{1}{c}{26.93} 
& \multicolumn{1}{c}{21.03} 
& \multicolumn{1}{c}{18.33}\\ 

\multicolumn{1}{l|}{(b)}
& \checkmark
&
&
& \multicolumn{1}{c}{21.35} 
& \multicolumn{1}{c}{16.97} 
& \multicolumn{1}{c|}{14.59}
& \multicolumn{1}{c}{29.54} 
& \multicolumn{1}{c}{22.71} 
& \multicolumn{1}{c}{19.69}\\

\multicolumn{1}{l|}{(c)}
& 
& \checkmark
&
& \multicolumn{1}{c}{21.26} 
& \multicolumn{1}{c}{16.93} 
& \multicolumn{1}{c|}{14.47}
& \multicolumn{1}{c}{29.16} 
& \multicolumn{1}{c}{22.44} 
& \multicolumn{1}{c}{19.53}\\

\multicolumn{1}{l|}{(d)}
& 
&
& \checkmark
& \multicolumn{1}{c}{22.23} 
& \multicolumn{1}{c}{17.60} 
& \multicolumn{1}{c|}{15.02}
& \multicolumn{1}{c}{31.47} 
& \multicolumn{1}{c}{23.75} 
& \multicolumn{1}{c}{21.46}\\

\multicolumn{1}{l|}{(e)}
& \checkmark
& \checkmark
&
& \multicolumn{1}{c}{21.41} 
& \multicolumn{1}{c}{17.17} 
& \multicolumn{1}{c|}{14.67}
& \multicolumn{1}{c}{29.71} 
& \multicolumn{1}{c}{22.97} 
& \multicolumn{1}{c}{20.04}\\

\multicolumn{1}{l|}{(f)}
& \checkmark
& 
& \checkmark
& \multicolumn{1}{c}{24.74} 
& \multicolumn{1}{c}{18.44} 
& \multicolumn{1}{c|}{15.63}
& \multicolumn{1}{c}{33.83} 
& \multicolumn{1}{c}{25.18} 
& \multicolumn{1}{c}{21.90}\\

\multicolumn{1}{l|}{(g)}
& 
& \checkmark
& \checkmark
& \multicolumn{1}{c}{24.63} 
& \multicolumn{1}{c}{18.32} 
& \multicolumn{1}{c|}{15.49}
& \multicolumn{1}{c}{33.30} 
& \multicolumn{1}{c}{25.14} 
& \multicolumn{1}{c}{21.82}\\

\multicolumn{1}{l|}{(h)}
& \checkmark
& \checkmark
& \checkmark
& \multicolumn{1}{c}{\textbf{26.10}}
& \multicolumn{1}{c}{\textbf{19.18}}
& \multicolumn{1}{c|}{\textbf{16.96}}
& \multicolumn{1}{c}{\textbf{34.77}} 
& \multicolumn{1}{c}{\textbf{25.75}} 
& \multicolumn{1}{c}{\textbf{22.44}}\\

\midrule
\multicolumn{1}{l|}{}
&&&
& \multicolumn{1}{c}{+6.24} 
& \multicolumn{1}{c}{+4.07} 
& \multicolumn{1}{c|}{+4.32}
& \multicolumn{1}{c}{+7.84} 
& \multicolumn{1}{c}{+4.72} 
& \multicolumn{1}{c}{+4.11}\\

\bottomrule
\end{tabular}
\label{table_kitti_val_ab}
\end{table}

\begin{figure*}[t]
\centering {\includegraphics[width=0.323\textwidth]{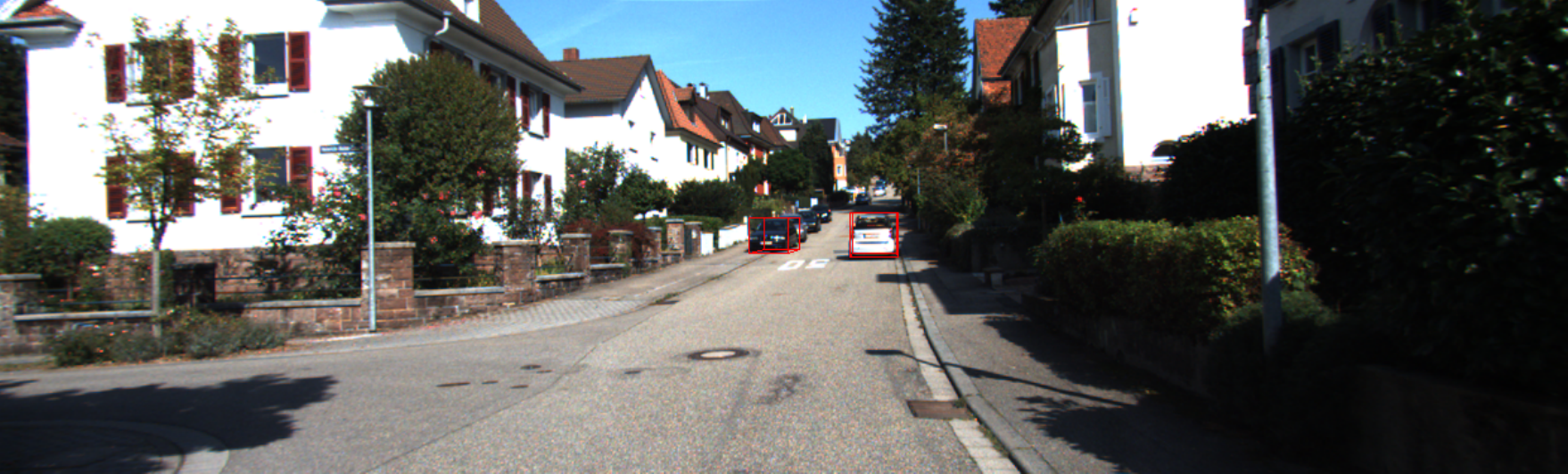}}
\centering {\includegraphics[width=0.323\textwidth]{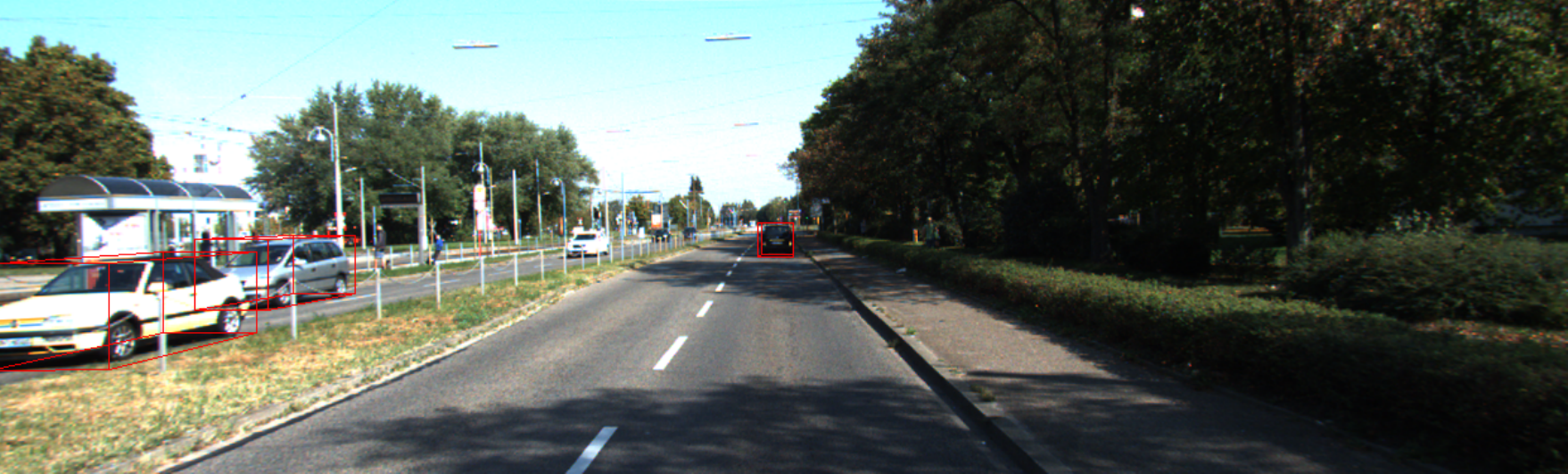}}
\centering {\includegraphics[width=0.323\textwidth]{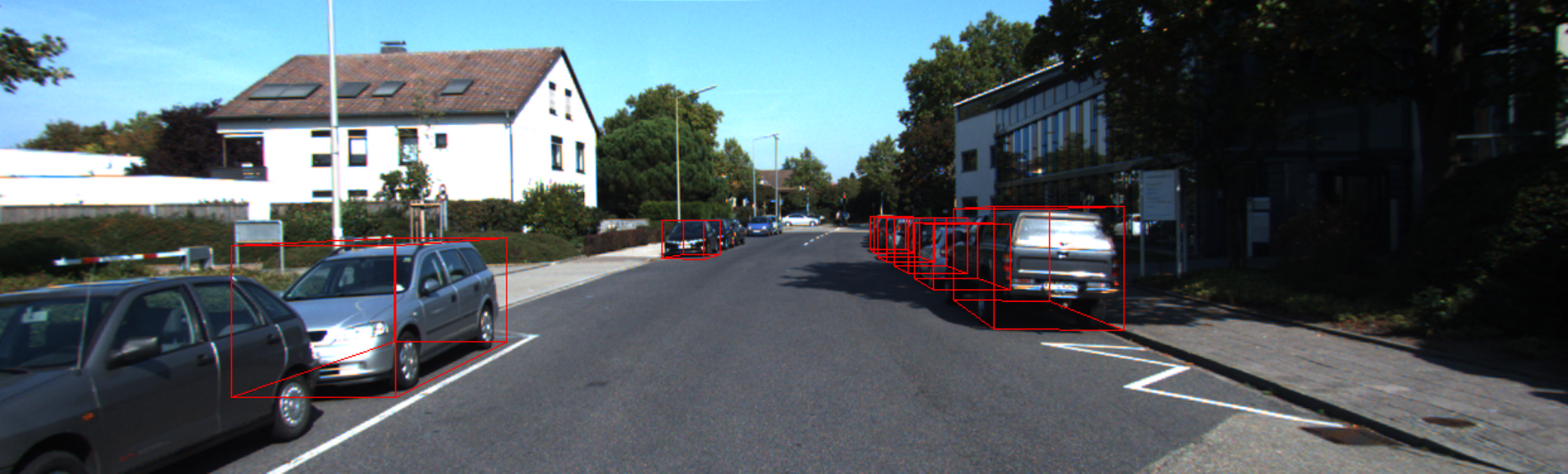}}
\\
\centering {\includegraphics[width=0.323\textwidth]{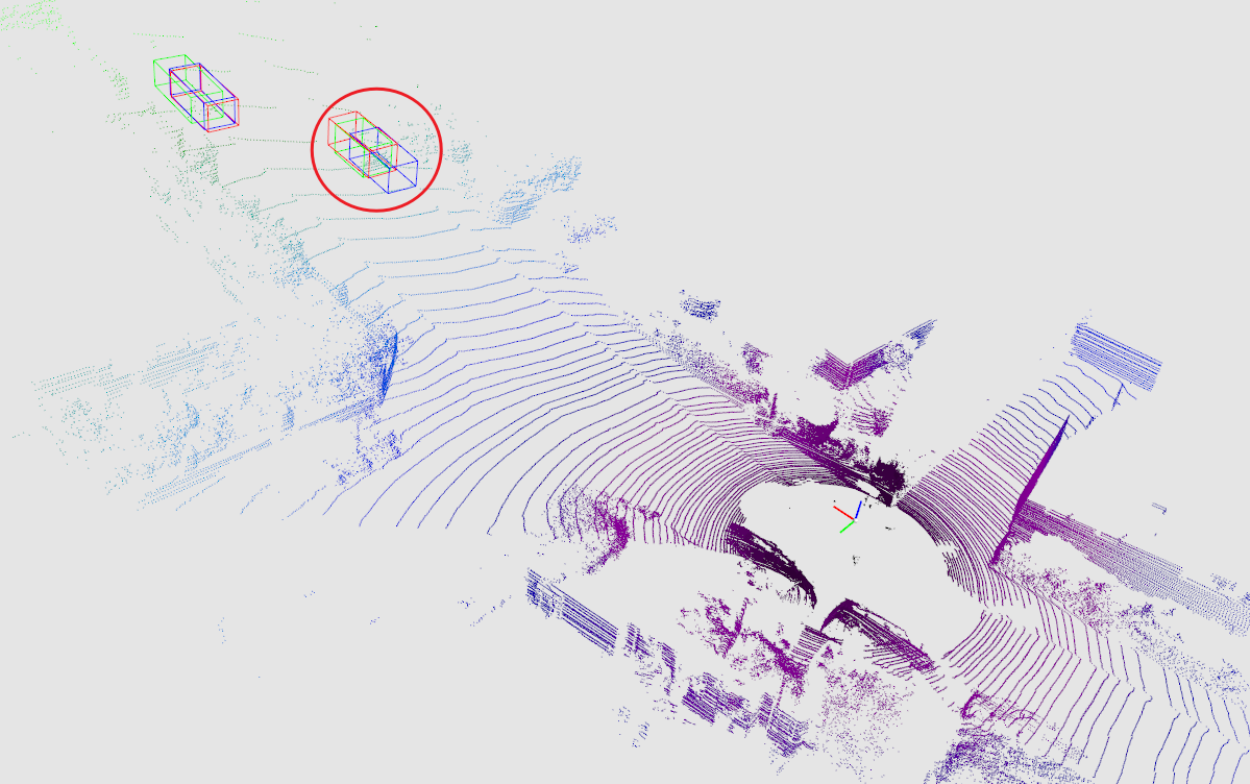}}
\centering {\includegraphics[width=0.323\textwidth]{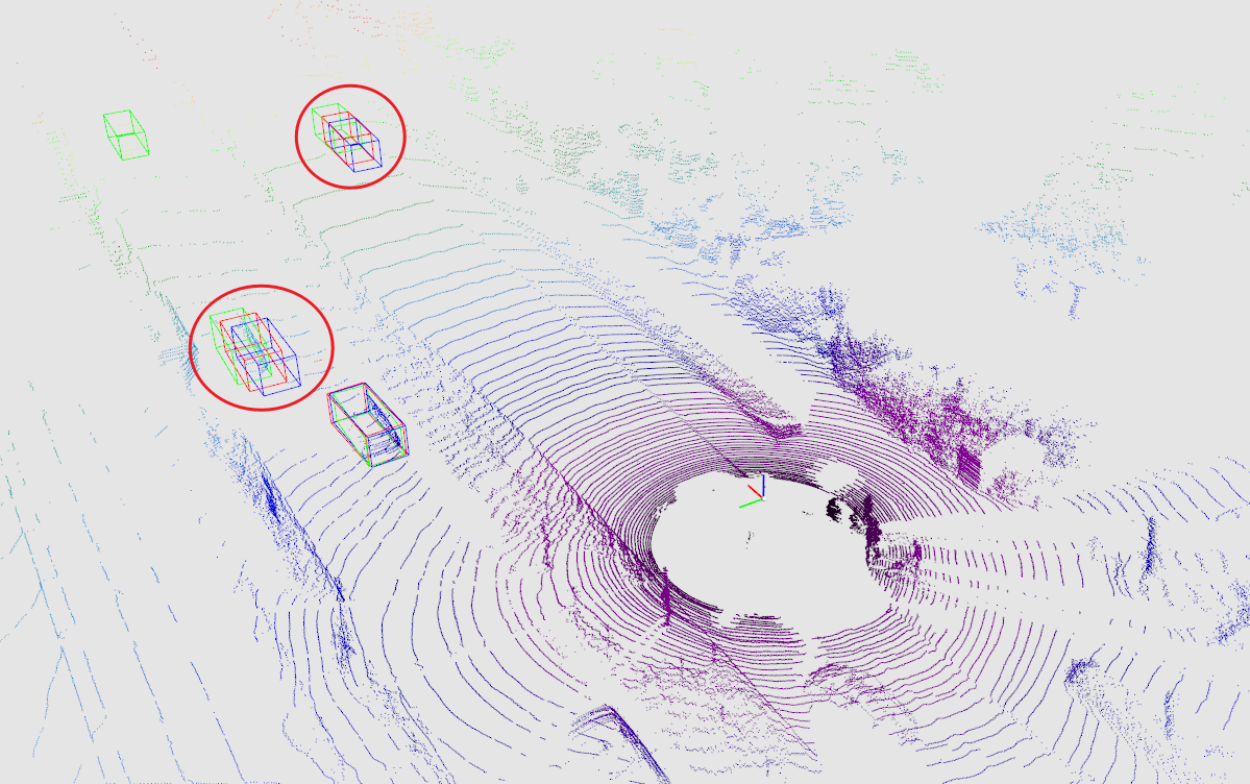}}
\centering {\includegraphics[width=0.323\textwidth]{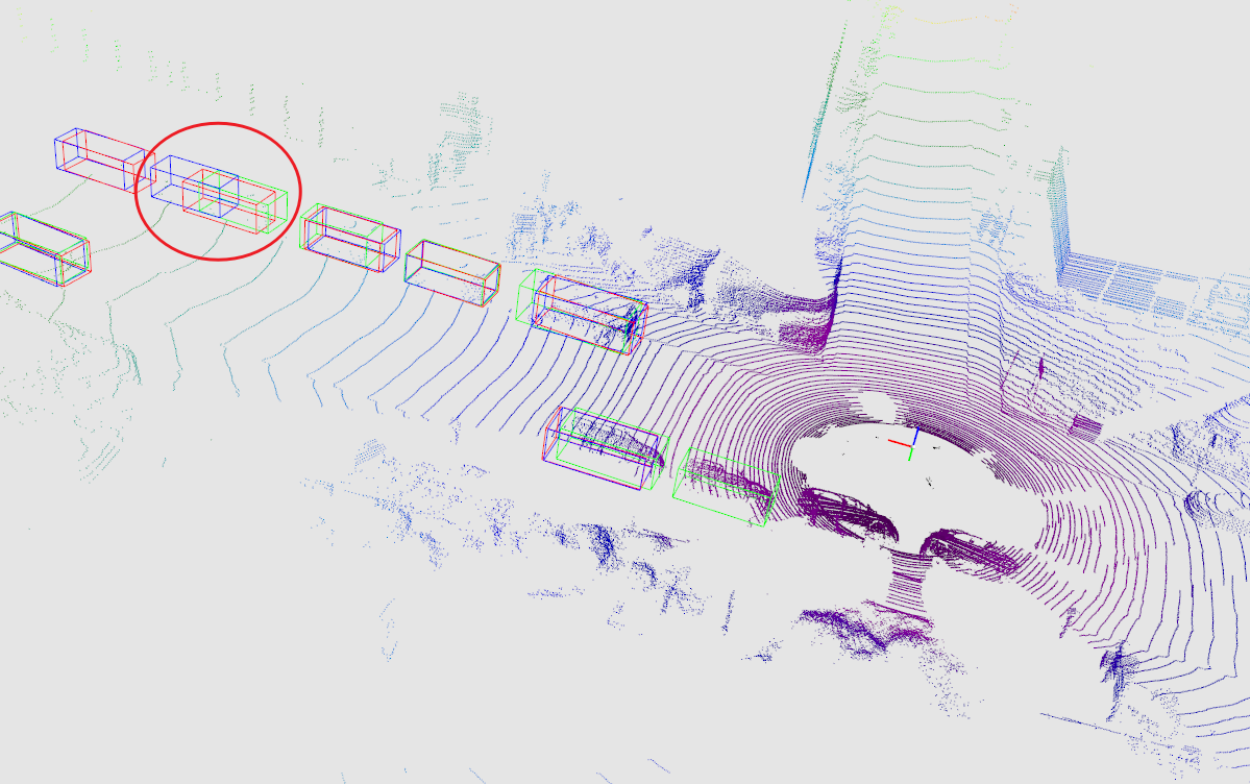}}
\caption{
\textbf{Qualitative results.}
We employ blue, red, and green 3D boxes to denote the DID-M3D baseline, MonoSKD, and ground truth results. Additionally, we use red circles to highlight significant differences.
}
\label{Qualitative results}
\end{figure*}


\subsection{State-of-the-art Comparisions}
As is shown in Table \ref{table_kitti_test}, we compare the experimental results of our framework and other state-of-the-art methods on the KITTI \textit{testing} set.
Our schemes are significantly improved compared to the respective baseline models and outperform other state-of-the-art methods.
On the car category that KITTI cares most about, our MonoSKD+DID-M3D results achieve state-of-the-art performance on almost all 3D-level and BEV-level metrics.
In particular, compared with the second-best results, our scheme achieves up to 13.31\% and 7.81\% relative performance improvements on 3D-level and BEV-level metrics.
Our scheme can theoretically be applied to any monocular 3D detector and improve performance.

\subsection{Ablation Study}
In this section, we analyze the effectiveness of each part of our distillation framework on the KITTI \textit{validation} set.
As shown in Table \ref{table_kitti_val_ab}, considering that GUPNet and DID-M3D uses the ROI-align operation and the reproducibility cannot be guaranteed, we conduct the ablation experiment base on the MonoDLE network.
It is noteworthy that we conduct our ablation experiments on the redesigned distillation framework (discussed in Section \ref{Selection of Distillation Objects}). 
Our redesigned distillation framework is higher than the original MonoDistill in most indicators, which has lower GPU memory occupation and faster training speed (see Appendix \ref{GPU}).
Specifically, all distillation schemes improve the accuracy of the baseline model, and their improvements are complementary.
Compared with the baseline, the final model can improve the \textbf{absolute} detection performance by \textbf{6.24}, \textbf{4.07}, \textbf{4.32} and \textbf{absolute} BEV performance by \textbf{7.84}, \textbf{4.72}, \textbf{4.11} on the easy, moderate, and hard levels, respectively.



\subsection{Distillation with Different Detectors and Backbones}
To further demonstrate the effectiveness and generalizability of our method, we select three monocular 3D detection networks and compare our method with the competitive MonoDistill scheme.
As shown in Table \ref{table_more_detector}, our distillation scheme has achieved better performance than MonoDistill on all detectors in terms of detection performance and BEV performance, which also verifies the superiority of our scheme.

In Table \ref{table_more_backbone}, we select representative backbones like ResNet~\cite{ResNet} and MobileNetv3~\cite{mobilenetv3} 
to supplement related experiments, and the results show that our scheme is not sensitive to the backbones.

\begin{table}
\vspace{-5pt}
\centering
\caption{
\textbf{Distillation performance with different detectors of the Car category on the KITTI \textit{validation} set.}
${\dagger}$: our reproduced results, whose performance is much higher than that reported in the original MonoDistill paper.
}
\begin{tabular}{c|c|c|c|c|c|c|c}
\toprule
    \multicolumn{2}{c|}{\multirow{2}{*}{Methods}}
    &\multicolumn{3}{c|}{$AP_{3D}$ (Car val) }
    &\multicolumn{3}{c}{$AP_{BEV}$ (Car val) }\\

\multicolumn{2}{c|}{}
    & \multicolumn{1}{c}{Easy} 
    & \multicolumn{1}{c}{Mod.} 
    & \multicolumn{1}{c|}{Hard}
    & \multicolumn{1}{c}{Easy} 
    & \multicolumn{1}{c}{Mod.} 
    & \multicolumn{1}{c}{Hard}\\
    
\midrule

\multicolumn{2}{l|}{MonoDLE}
    & \multicolumn{1}{c}{19.86} 
& \multicolumn{1}{c}{15.11} 
& \multicolumn{1}{c|}{12.64}
& \multicolumn{1}{c}{26.93} 
& \multicolumn{1}{c}{21.03} 
& \multicolumn{1}{c}{18.33}\\ 
    

\multicolumn{2}{l|}{+MonoDistill}
    & \multicolumn{1}{c}{24.40} 
    & \multicolumn{1}{c}{18.47} 
    & \multicolumn{1}{c|}{16.46}
    & \multicolumn{1}{c}{32.86} 
    & \multicolumn{1}{c}{25.14}
    & \multicolumn{1}{c}{21.99}\\
    

\multicolumn{2}{l|}{+Ours}
    & \multicolumn{1}{c}{\textbf{26.10}} 
    & \multicolumn{1}{c}{\textbf{19.18}} 
    & \multicolumn{1}{c|}{\textbf{16.96}}
    & \multicolumn{1}{c}{\textbf{34.77}}
    & \multicolumn{1}{c}{\textbf{25.75}}
    & \multicolumn{1}{c}{\textbf{22.44}}\\

\midrule
\multicolumn{2}{l|}{GUPNet}
    & \multicolumn{1}{c}{21.19} 
    & \multicolumn{1}{c}{16.23} 
    & \multicolumn{1}{c|}{13.57}
    & \multicolumn{1}{c}{30.14} 
    & \multicolumn{1}{c}{22.38}
    & \multicolumn{1}{c}{19.29}\\
    

\multicolumn{2}{l|}{+MonoDistill$^{\dagger}$}

    & \multicolumn{1}{c}{24.34} 
    & \multicolumn{1}{c}{17.72} 
    & \multicolumn{1}{c|}{14.89}
    & \multicolumn{1}{c}{31.74} 
    & \multicolumn{1}{c}{23.22}
    & \multicolumn{1}{c}{19.98}\\
    

\multicolumn{2}{l|}{+Ours}
    & \multicolumn{1}{c}{\textbf{25.30}} 
    & \multicolumn{1}{c}{\textbf{18.06}} 
    & \multicolumn{1}{c|}{\textbf{15.37}}
    & \multicolumn{1}{c}{\textbf{32.54}}
    & \multicolumn{1}{c}{\textbf{23.72}}
    & \multicolumn{1}{c}{\textbf{21.71}}\\
    

\midrule
\multicolumn{2}{l|}{DID-M3D}
    & \multicolumn{1}{c}{25.75} 
    & \multicolumn{1}{c}{17.77} 
    & \multicolumn{1}{c|}{14.74}
    & \multicolumn{1}{c}{33.39} 
    & \multicolumn{1}{c}{23.66} 
    & \multicolumn{1}{c}{20.86}\\

\multicolumn{2}{l|}{+MonoDistill}
    & \multicolumn{1}{c}{27.08} 
    & \multicolumn{1}{c}{19.31} 
    & \multicolumn{1}{c|}{16.16}
    & \multicolumn{1}{c}{35.85} 
    & \multicolumn{1}{c}{25.47}
    & \multicolumn{1}{c}{21.73}\\

\multicolumn{2}{l|}{+Ours}
    & \multicolumn{1}{c}{\textbf{28.91}} 
    & \multicolumn{1}{c}{\textbf{20.21}} 
    & \multicolumn{1}{c|}{\textbf{16.99}}
    & \multicolumn{1}{c}{\textbf{37.66}}
    & \multicolumn{1}{c}{\textbf{26.41}}
    & \multicolumn{1}{c}{\textbf{23.39}}\\

\bottomrule
\end{tabular}
\label{table_more_detector}
\vspace{-15pt}
\end{table}

\begin{table}
\setlength\tabcolsep{5.4pt}
\centering
\caption{
    \textbf{Quantitative comparison of different backbones on KITTI \textit{validation} set.}
    We conduct experiments based on MonoDLE network.
    The best results are listed in \textbf{bold}.
}
\begin{tabular}{c|c|c|c|c|c|c|c|c}
\toprule
    \multicolumn{2}{c|}{\multirow{2}{*}{Backbones}}
    &\multicolumn{3}{c|}{$AP_{3D}$ (Car val)}
    &\multicolumn{3}{c}{$AP_{BEV}$ (Car val)}\\

\multicolumn{2}{c|}{} 
    & \multicolumn{1}{c}{Easy} 
    & \multicolumn{1}{c}{Mod.} 
    & \multicolumn{1}{c|}{Hard} 
    & \multicolumn{1}{c}{Easy} 
    & \multicolumn{1}{c}{Mod.}
    & \multicolumn{1}{c}{Hard} \\
\midrule

\multicolumn{2}{l|}{\multirow{1}{*}{Res18 (T)}}
    & \multicolumn{1}{c}{66.54}
    & \multicolumn{1}{c}{48.31} 
    & \multicolumn{1}{c|}{41.56}
    & \multicolumn{1}{c}{77.78} 
    & \multicolumn{1}{c}{61.60}
    & \multicolumn{1}{c}{52.96} \\
\multicolumn{2}{l|}{Res18 (S)}
    & \multicolumn{1}{c}{17.96} 
    & \multicolumn{1}{c}{13.91} 
    & \multicolumn{1}{c|}{12.17}
    & \multicolumn{1}{c}{24.78} 
    & \multicolumn{1}{c}{18.74} 
    & \multicolumn{1}{c}{16.72}\\
\multicolumn{2}{l|}{+MonoDistill}
    & \multicolumn{1}{c}{20.25} 
    & \multicolumn{1}{c}{15.60} 
    & \multicolumn{1}{c|}{13.12}
    & \multicolumn{1}{c}{27.88} 
    & \multicolumn{1}{c}{20.83} 
    & \multicolumn{1}{c}{17.95}\\
\multicolumn{2}{l|}{+MonoSKD}
    & \multicolumn{1}{c}{\textbf{21.58}}
    & \multicolumn{1}{c}{\textbf{16.31}}
    & \multicolumn{1}{c|}{\textbf{13.70}}
    & \multicolumn{1}{c}{\textbf{28.52}} 
    & \multicolumn{1}{c}{\textbf{21.55}} 
    & \multicolumn{1}{c}{\textbf{18.48}} \\
\midrule
\multicolumn{2}{l|}{\multirow{1}{*}{MobileNetv3-L (T)}}
    & \multicolumn{1}{c}{65.77} 
    & \multicolumn{1}{c}{48.55} 
    & \multicolumn{1}{c|}{40.81}
    & \multicolumn{1}{c}{77.36} 
    & \multicolumn{1}{c}{60.75}
    & \multicolumn{1}{c}{52.19} \\
\multicolumn{2}{l|}{MobileNetv3-L (S)}
    & \multicolumn{1}{c}{15.73} 
    & \multicolumn{1}{c}{12.37} 
    & \multicolumn{1}{c|}{10.33}
    & \multicolumn{1}{c}{23.47} 
    & \multicolumn{1}{c}{17.57}
    & \multicolumn{1}{c}{15.67} \\
\multicolumn{2}{l|}{+MonoDistill}
    & \multicolumn{1}{c}{16.59} 
    & \multicolumn{1}{c}{12.93} 
    & \multicolumn{1}{c|}{10.71}
    & \multicolumn{1}{c}{25.34} 
    & \multicolumn{1}{c}{19.58} 
    & \multicolumn{1}{c}{16.81}\\
\multicolumn{2}{l|}{+MonoSKD}
    & \multicolumn{1}{c}{\textbf{18.15}}
    & \multicolumn{1}{c}{\textbf{14.02}}
    & \multicolumn{1}{c|}{\textbf{12.28}}
    & \multicolumn{1}{c}{\textbf{26.97}} 
    & \multicolumn{1}{c}{\textbf{20.24}} 
    & \multicolumn{1}{c}{\textbf{17.34}} \\

\bottomrule
\end{tabular}
\label{table_more_backbone}
\end{table}

\subsection{Qualitative Results}
In order to demonstrate the superiority of our method more intuitively, we visualize the results predicted by the network. 
As shown in Figure \ref{Qualitative results}, we apply the red boxes to represent the result of our proposed MonoSKD+DID-M3D. 
Our model has better localization performance than the baseline model.

\subsection{Pedestrian/Cyclist Detection}
To demonstrate the generalizability of other categories, we perform experiments on cyclist and pedestrian categories (see Table \ref{table_pedestrian_cyclist}).
We use the pre-trained model provided by DID-M3D as the baseline and retrain the three-category teacher model.
Our scheme achieves the best results in almost all indicators.
Moreover, we provide the full three-category performance in Appendix Table \ref{Appendix_ped_cyc_car}.
It is worth noting that the performance of the \textbf{Cyclist} in MonoDistill actually drops after distillation.
We guess this is because the bounding boxes of the Cyclist category often contain more background pixels, so the alignment based on L1 loss misleads learning.
In contrast, our Spearman distillation scheme is more relaxed and steadily improves detection performance.

\begin{table}
\centering
\vspace{-5pt}
\caption{
\textbf{Performance of Pedestrian/Cyclist detection on the KITTI \textit{validation} set under
IoU criterion 0.5.}
The best results are listed in bold.
}
\label{table_pedestrian_cyclist}
\begin{tabular}{c|c|c|c|c|c|c|c}
\toprule
    \multicolumn{2}{c|}{\multirow{2}{*}{Methods}}
    &\multicolumn{3}{c|}{$AP_{3D}$(Pedestrian)}
    &\multicolumn{3}{c}{$AP_{3D}$(Cyclist)}\\

\multicolumn{2}{c|}{}
    & \multicolumn{1}{c}{Easy} 
    & \multicolumn{1}{c}{Mod.} 
    & \multicolumn{1}{c|}{Hard}
    & \multicolumn{1}{c}{Easy} 
    & \multicolumn{1}{c}{Mod.} 
    & \multicolumn{1}{c}{Hard}\\
\midrule
\multicolumn{2}{l|}{DID-M3D}
    & \multicolumn{1}{c}{11.15} 
    & \multicolumn{1}{c}{8.65} 
    & \multicolumn{1}{c|}{7.15}
    & \multicolumn{1}{c}{5.40} 
    & \multicolumn{1}{c}{2.81} 
    & \multicolumn{1}{c}{2.72}\\
\multicolumn{2}{l|}{+MonoDistill}
    & \multicolumn{1}{c}{15.96} 
    & \multicolumn{1}{c}{11.86} 
    & \multicolumn{1}{c|}{\textbf{9.80}}
    & \multicolumn{1}{c}{4.73} 
    & \multicolumn{1}{c}{2.69} 
    & \multicolumn{1}{c}{2.50}\\
\multicolumn{2}{l|}{+MonoSKD}
    & \multicolumn{1}{c}{\textbf{16.24}} 
    & \multicolumn{1}{c}{\textbf{11.92}} 
    & \multicolumn{1}{c|}{9.28}
    & \multicolumn{1}{c}{\textbf{5.50}}
    & \multicolumn{1}{c}{\textbf{3.45}}
    & \multicolumn{1}{c}{\textbf{3.00}}\\

\bottomrule
\end{tabular}
\vspace{-10pt}
\end{table}

\subsection{Sensitivity study of loss weight $\alpha$}
In Eq. \ref{eq6}, we use the loss weight hyper-parameter $\alpha$ to balance the detection training loss and distillation loss.
Here, we conduct several experiments to investigate the influence of $\alpha$.
As shown in Table \ref{ablation_alpha}, the worst result is just a 0.38 mAP drop compared with the best result (19.18 $\rightarrow$ 18.80), indicating our method is not sensitive to the hyper-parameter $\alpha$.

\begin{table}
\centering
\vspace{-5pt}
\caption{
\textbf{Ablation study of loss weight hyper-parameter $\alpha$.}}
\begin{tabular}{c|c|c|c|c|c|c|c|c|c}
\toprule
    \multicolumn{2}{c|}{\multirow{2}{*}{$\alpha$}}
    &\multicolumn{3}{c|}{$AP_{3D}$ (Car val)}
    &\multicolumn{3}{c}{$AP_{BEV}$ (Car val)}\\
    \multicolumn{2}{c|}{} 
    & \multicolumn{1}{c}{Easy} 
    & \multicolumn{1}{c}{Mod.} 
    & \multicolumn{1}{c|}{Hard}
    & \multicolumn{1}{c}{Easy} 
    & \multicolumn{1}{c}{Mod.}
    & \multicolumn{1}{c}{Hard} \\
    
    \midrule
    \multicolumn{2}{l|}{\multirow{1}{*}{0.5}}
    & \multicolumn{1}{c}{25.10}
    & \multicolumn{1}{c}{18.80} 
    & \multicolumn{1}{c|}{16.58}
    & \multicolumn{1}{c}{34.42} 
    & \multicolumn{1}{c}{25.60}
    & \multicolumn{1}{c}{22.16} \\
    \multicolumn{2}{l|}{\multirow{1}{*}{1.0}}
    & \multicolumn{1}{c}{26.10}
    & \multicolumn{1}{c}{\textbf{19.18}} 
    & \multicolumn{1}{c|}{\textbf{16.96}}
    & \multicolumn{1}{c}{34.77} 
    & \multicolumn{1}{c}{\textbf{25.75}}
    & \multicolumn{1}{c}{\textbf{22.44}} \\ 
    \multicolumn{2}{l|}{\multirow{1}{*}{2.0}}
    & \multicolumn{1}{c}{\textbf{26.38}}
    & \multicolumn{1}{c}{19.05} 
    & \multicolumn{1}{c|}{16.81}
    & \multicolumn{1}{c}{\textbf{34.81}} 
    & \multicolumn{1}{c}{25.68}
    & \multicolumn{1}{c}{22.38} \\
\bottomrule
\end{tabular}
\label{ablation_alpha}
\vspace{-15pt}
\end{table}

\subsection{Insights into Spearman Distillation and Downstream Task Generality}

Spearman distillation introduces a loose constraint between cross-modal features instead of strict alignment and is more suitable for cross-modal tasks.
The 3D object detection task is a typical cross-model task for its multiple modalities (e.g., Camera, LiDAR, and Radar) inputs, and thus, we choose monocular 3D
object detection task to verify our method.
We can put the whole story on the knowledge distillation itself.
Spearman distillation has strong knowledge transfer potential and can be easily extended to downstream tasks.
Take 2D detection as an example (Table \ref{coco_retina}). 
SKD can achieve significant performance gains without scene relation distillation, even with \textbf{heterogeneous} teachers.
CMKD is a distillation method for BEV paradigm 3D detection, so we replaced the MSE loss in CMKD with the proposed SKD.
Table \ref{bev} reveals that SKD suits the BEV paradigm detectors. 
Compared with CMKD, SKD brings performance improvements in almost all indicators.

\begin{table}
\vspace{-5pt}
\centering
\caption{\textbf{Distilling Student Detectors with Homogeneous and Heterogeneous Teachers on the COCO dataset.}}
\label{coco}
    \begin{tabular}{cccccc}
        \toprule
        Method          & schedule  & $mAP$    & $AP_{S}$    & $AP_{M}$    & $AP_{L}$ \\
        \midrule
        Retina-ResX101 (T)  & 2x    & 40.8  & 22.9   & 44.5   & 54.6 \\
        Retina-Res50 (S)    & 2x    & 37.4  & 20.0   & 40.7   & 49.7 \\
        +FRS\cite{FRS}       & 2x    & 40.1  & 21.9   & 43.7   & 54.3 \\
        +FGD\cite{FGD}       & 2x    & 40.4  & \textbf{23.4}   & 44.7   & 54.1 \\
        +SKD (Ours)          & 2x    
                            & \textbf{40.6}
                            & 22.0  
                            & \textbf{44.8}      
                            & \textbf{54.8} \\
        \midrule
        FCOS-X101 (T)    & 2x+ms    & 42.7  & 26.0      & 46.5      & 54.7 \\
        Retina-Res50 (S) & 1x       & 36.5  & 20.4      & 40.3      & 48.1 \\
        +SKD (Ours)       & 1x       & \textbf{40.1}  & \textbf{22.7}      & \textbf{44.3}      & \textbf{53.8} \\
        \bottomrule
    \end{tabular}
\label{coco_retina}
\vspace{-5pt}
\end{table}


\begin{table}
\setlength\tabcolsep{4.5pt}
\centering
\caption{\textbf{Effectiveness experiments under the BEV paradigm.}}
\label{bev}
\begin{tabular}{c|c|c|c|c|c|c|c}
\toprule
    \multicolumn{2}{c|}{\multirow{2}{*}{Method}}
    &\multicolumn{3}{c|}{$AP_{3D}$ (Car val)}
    &\multicolumn{3}{c}{$AP_{BEV}$ (Car val)}\\
    
    \multicolumn{2}{c|}{}
    & \multicolumn{1}{c}{Easy} 
    & \multicolumn{1}{c}{Mod.} 
    & \multicolumn{1}{c|}{Hard}
    & \multicolumn{1}{c}{Easy} 
    & \multicolumn{1}{c}{Mod.} 
    & \multicolumn{1}{c}{Hard}\\
    
    \midrule
    \multicolumn{2}{l|}{CMKD\cite{CMKD}} 
    & \multicolumn{1}{c}{23.49} 
    & \multicolumn{1}{c}{15.56} 
    & \multicolumn{1}{c|}{12.97}
    & \multicolumn{1}{c}{\textbf{31.81}} 
    & \multicolumn{1}{c}{21.02}
    & \multicolumn{1}{c}{18.57}\\
    
    \multicolumn{2}{l|}{+SKD (Ours)} 
    & \multicolumn{1}{c}{\textbf{23.59}} 
    & \multicolumn{1}{c}{\textbf{15.79}} 
    & \multicolumn{1}{c|}{\textbf{13.16}}
    & \multicolumn{1}{c}{31.79} 
    & \multicolumn{1}{c}{\textbf{22.27}} 
    & \multicolumn{1}{c}{\textbf{19.05}}\\
    \midrule
\end{tabular}
\vspace{-10pt}
\end{table}

\section{Conclusion}
In this paper, we propose \textbf{MonoSKD}, a cross-modal distillation framework for monocular 3D object detection via Spearman’s rank correlation coefficient.
Existing distillation schemes try to strictly align cross-modal features, thus leading to suboptimal distillation performance.
To alleviate this problem, we propose to use the Spearman correlation coefficient to help mine ranking knowledge among features.
To improve distillation efficiency, we select the appropriate distillation objects to save 30\% of GPU memory and accelerate training. 
We distill three detectors to verify the effectiveness of our scheme and achieve state-of-the-art performance on the challenging KITTI benchmark without introducing additional inference costs.


\bibliography{ecai}

\clearpage
\appendix
\section{Appendix}
Considering the space constraints of the main text, we provide more experimental results and discussions in the supplementary material.

\subsection{Motivation}
In recent years, remarkable progress has been made in monocular 3D object detection.
However, these lightweight detectors face the problem of low detection performance, so distillation frameworks such as MonoDistill are proposed to alleviate this problem.
In our research on MonoDistill, we find that the distillation technique used by MonoDistill is a stricter constraint based on the 2D detection distillation scheme.
Because 2D object detection uses RGB image input, the difference between the teacher and student models is not vast, so that we can use a strict distillation strategy. 
When it comes to monocular 3D object detection, there is a vast difference in the input of the teacher and the student model. Directly aligning features may mislead the training, so finding a general and loose distillation strategy is necessary.
Specifically, we try to distill the relative relationship between features, introducing the Spearman correlation coefficient.

 


\begin{table}
\centering
\caption{
\textbf{Quantitative comparison between teacher and student on the KITTI \textit{validation} set.}
'T' indicates the teacher using the dense depth maps as input for training and inference.
'S' indicates the student without distillation.
}
\begin{tabular}{c|c|c|c|c|c|c|c|c}
\toprule
    \multicolumn{2}{c|}{\multirow{2}{*}{Methods}}
    &\multicolumn{3}{c|}{$AP_{3D}$ (Car val)}
    &\multicolumn{3}{c}{$AP_{BEV}$ (Car val)}\\

\multicolumn{2}{c|}{} 
    & \multicolumn{1}{c}{Easy} 
    & \multicolumn{1}{c}{Mod.} 
    & \multicolumn{1}{c|}{Hard} 
    & \multicolumn{1}{c}{Easy} 
    & \multicolumn{1}{c}{Mod.}
    & \multicolumn{1}{c}{Hard} \\
\midrule

\multicolumn{2}{l|}{\multirow{1}{*}{MonoDLE (T)}}
    & \multicolumn{1}{c}{60.57}
    & \multicolumn{1}{c}{45.06} 
    & \multicolumn{1}{c|}{37.90}
    & \multicolumn{1}{c}{74.20} 
    & \multicolumn{1}{c}{57.49}
    & \multicolumn{1}{c}{50.89} \\
\multicolumn{2}{l|}{MonoDLE (S)}
    & \multicolumn{1}{c}{19.86} 
    & \multicolumn{1}{c}{15.11} 
    & \multicolumn{1}{c|}{12.64}
    & \multicolumn{1}{c}{26.93} 
    & \multicolumn{1}{c}{21.03} 
    & \multicolumn{1}{c}{18.33}\\ 
\midrule
\multicolumn{2}{l|}{\multirow{1}{*}{GUPNet (T)}} 
    & \multicolumn{1}{c}{48.54} 
    & \multicolumn{1}{c}{32.88} 
    & \multicolumn{1}{c|}{27.44}
    & \multicolumn{1}{c}{61.29} 
    & \multicolumn{1}{c}{43.58}
    & \multicolumn{1}{c}{37.63} \\
\multicolumn{2}{l|}{GUPNet (S)}
    & \multicolumn{1}{c}{21.19} 
    & \multicolumn{1}{c}{16.23} 
    & \multicolumn{1}{c|}{13.57}
    & \multicolumn{1}{c}{30.14} 
    & \multicolumn{1}{c}{22.38}
    & \multicolumn{1}{c}{19.29} \\
\midrule
\multicolumn{2}{l|}{\multirow{1}{*}{DID-M3D (T)}}
    & \multicolumn{1}{c}{63.71}
    & \multicolumn{1}{c}{43.81} 
    & \multicolumn{1}{c|}{36.97}
    & \multicolumn{1}{c}{74.59} 
    & \multicolumn{1}{c}{55.09}
    & \multicolumn{1}{c}{46.52} \\
\multicolumn{2}{l|}{DID-M3D (S)}
    & \multicolumn{1}{c}{25.75} 
    & \multicolumn{1}{c}{17.77} 
    & \multicolumn{1}{c|}{14.74}
    & \multicolumn{1}{c}{33.39} 
    & \multicolumn{1}{c}{23.66}
    & \multicolumn{1}{c}{20.86} \\
\bottomrule
\end{tabular}
\label{table_stu_tea}
\end{table}

\subsection{More Details of Our Experiments}\label{More Details of Our Experiments}
To execute the distillation process, we have pre-trained the teacher network, and the performance of teacher and student is demonstrated in Table \ref{table_stu_tea}.

\textbf{MonoDLE.}
First, we chose MonoDLE as the baseline model for our ablation studies because we noticed that GUPNet and DID-M3D apply ROI-align operation, which is irreproducible during training.
We believe that this property affects the fairness of the experiment.
In contrast, the MonoDLE network has good reproducibility, so we do ablation experiments based on MonoDLE.
The same settings as MonoDistill are used when choosing MonoDLE as the baseline: learning rate, optimizer, and batch size.

\textbf{GUPNet.}
Because MonoDistill does not open-source the distillation code of GUPNet, we reproduce the relevant results, and our distillation results have higher performance.
In addition, the authors of GUPNet open-source the pre-trained model, so we directly reuse it as a student model.
Because we use a different PyTorch version, it is reasonable that the performance of the pre-trained models is slightly different.
For GUPNet, we use three categories in KITTI for training.

\textbf{DID-M3D.}
Following the original author's training setting, the DID-M3D results we report in the main text are all trained in the 'Car' category.
Again, we keep the same training settings as the original paper.
It is worth noting that when training the teacher models of MonoDLE and GUPNet, the GT depth map we use is provided by MonoDistill.
However, DID-M3D has already provided a pre-processed depth map. At this time, we directly reuse the depth map provided by DID-M3D.

\subsection{Advantages of the Redesigned Framework}\label{GPU}
\begin{figure*}[t]
\centering {\includegraphics[width=0.45\textwidth]{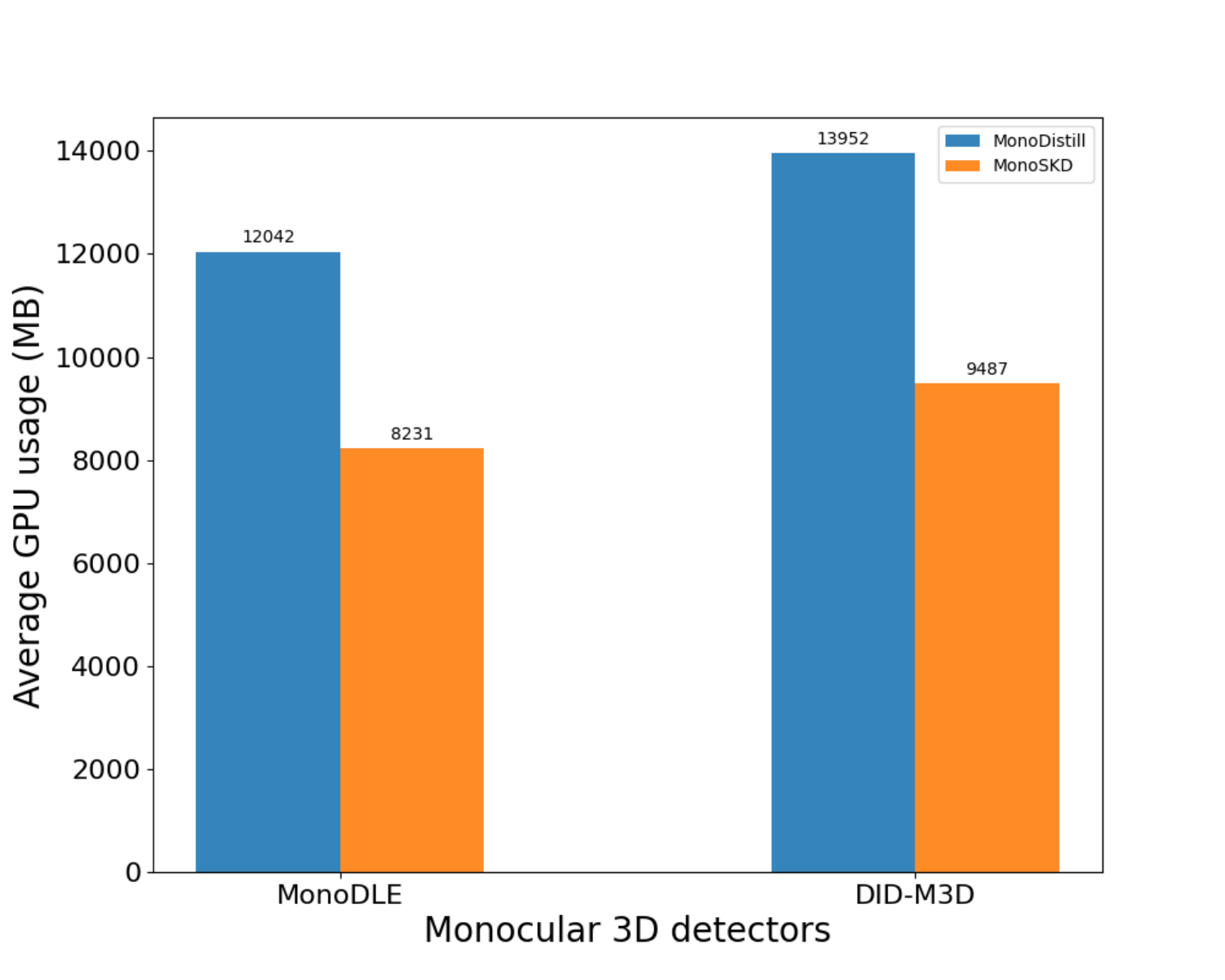}}
\centering {\includegraphics[width=0.45\textwidth]{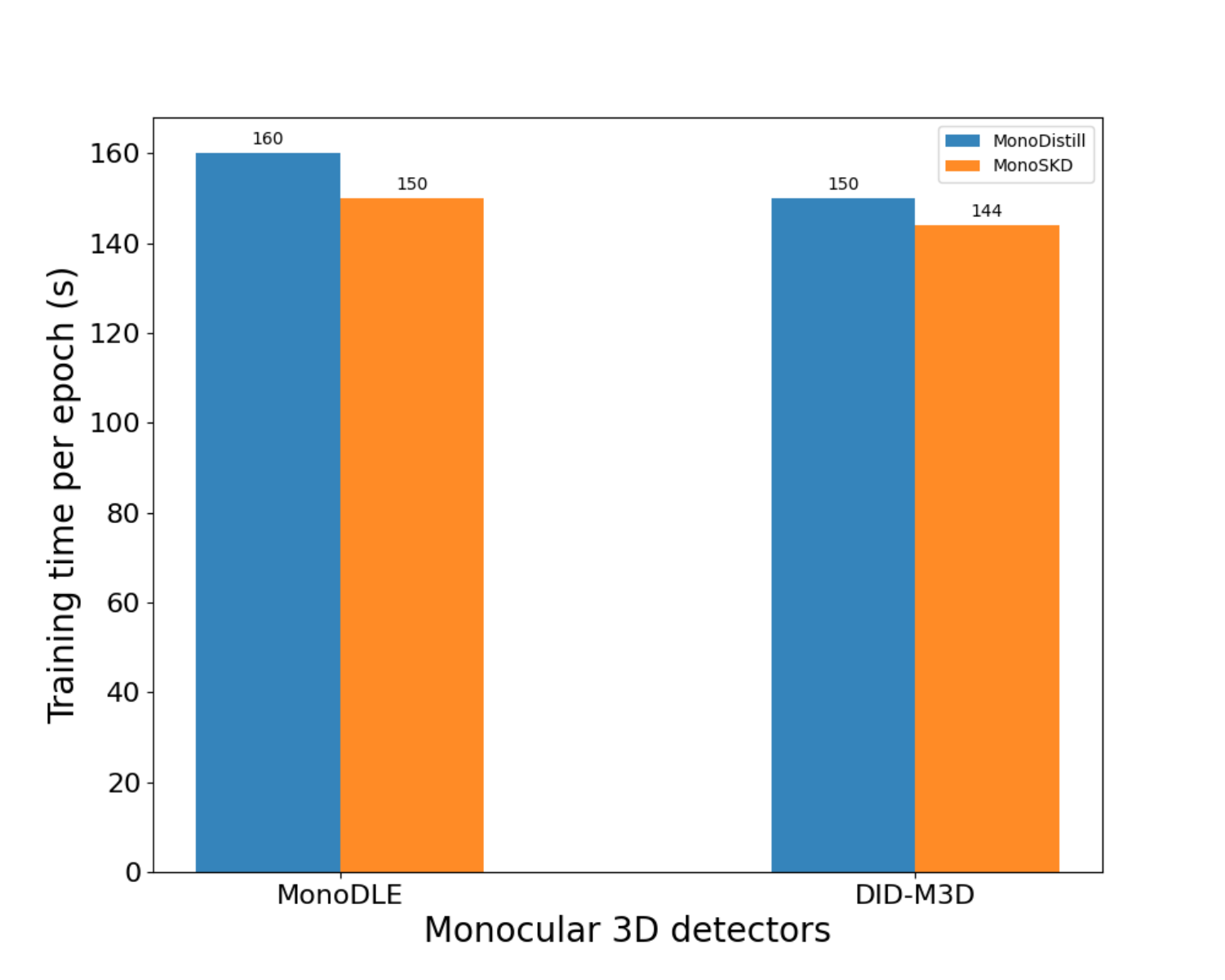}}
\caption{\textbf{Average GPU usage and training time of the redesigned distillation framework.} 
We use two GPUs, and the batch size of MonoDLE is 12, and the batch size of DID-M3D is 16.}
\label{table_gpu}
\end{figure*}

To validate the effectiveness of our distillation framework, we report the average GPU memory and training time per epoch for MonoDLE and DID-M3D, respectively.
As is shown in Figure \ref{table_gpu}, compared with MonoDistill, our distillation framework saves at least 30\% of GPU memory while bringing faster training.
In addition, we quantitatively show that the new framework still brings performance improvements in ablation experiments (see Table \ref{table_kitti_val_ab}).

\begin{table*}[htbp]
\setlength\tabcolsep{3pt}
\centering
\caption{
\textbf{Performance of Car/Pedestrian/Cyclist detection on the KITTI \textit{validation} set.}
Please note, Pedestrian/Cyclist performance is calculated under IoU criterion 0.5.
}
\label{Appendix_ped_cyc_car}
\begin{tabular}{c|c|c|c|c|c|c|c|c|c|c|c|c|c|c|c|c|c|c|c}
\hline
    \multicolumn{2}{c|}{\multirow{2}{*}{Methods}}
    &\multicolumn{3}{c|}{$AP_{3D}$ (Car)}
    &\multicolumn{3}{c|}{$AP_{BEV}$ (Car)}
    &\multicolumn{3}{c|}{$AP_{3D}$ (Ped.)}
    &\multicolumn{3}{c|}{$AP_{BEV}$ (Ped.)}
    &\multicolumn{3}{c|}{$AP_{3D}$ (Cyclist)}
    &\multicolumn{3}{c}{$AP_{BEV}$ (Cyclist)}\\

\multicolumn{2}{c|}{}
    & \multicolumn{1}{c}{Easy} 
    & \multicolumn{1}{c}{Mod.} 
    & \multicolumn{1}{c|}{Hard}
    & \multicolumn{1}{c}{Easy} 
    & \multicolumn{1}{c}{Mod.} 
    & \multicolumn{1}{c|}{Hard}
    & \multicolumn{1}{c}{Easy} 
    & \multicolumn{1}{c}{Mod.} 
    & \multicolumn{1}{c|}{Hard}
    & \multicolumn{1}{c}{Easy} 
    & \multicolumn{1}{c}{Mod.} 
    & \multicolumn{1}{c|}{Hard}
    & \multicolumn{1}{c}{Easy} 
    & \multicolumn{1}{c}{Mod.} 
    & \multicolumn{1}{c|}{Hard}
    & \multicolumn{1}{c}{Easy} 
    & \multicolumn{1}{c}{Mod.} 
    & \multicolumn{1}{c}{Hard}\\
\hline
\hline

\multicolumn{2}{l|}{DID-M3D}
    & \multicolumn{1}{c}{25.36} 
    & \multicolumn{1}{c}{17.03} 
    & \multicolumn{1}{c|}{14.05}
    & \multicolumn{1}{c}{33.90} 
    & \multicolumn{1}{c}{23.25} 
    & \multicolumn{1}{c|}{19.51} %
    & \multicolumn{1}{c}{11.15} 
    & \multicolumn{1}{c}{8.65} 
    & \multicolumn{1}{c|}{7.15}
    & \multicolumn{1}{c}{12.84} 
    & \multicolumn{1}{c}{10.23} 
    & \multicolumn{1}{c|}{7.90}
    & \multicolumn{1}{c}{5.40} 
    & \multicolumn{1}{c}{2.81} 
    & \multicolumn{1}{c|}{2.72}
    & \multicolumn{1}{c}{5.97} 
    & \multicolumn{1}{c}{3.50} 
    & \multicolumn{1}{c}{3.02}\\
    
\multicolumn{2}{l|}{+MonoDistill}
    & \multicolumn{1}{c}{26.14} 
    & \multicolumn{1}{c}{18.08} 
    & \multicolumn{1}{c|}{14.84}
    & \multicolumn{1}{c}{34.50} 
    & \multicolumn{1}{c}{24.56} 
    & \multicolumn{1}{c|}{20.84} %
    & \multicolumn{1}{c}{15.96} 
    & \multicolumn{1}{c}{11.86} 
    & \multicolumn{1}{c|}{\textbf{9.80}}
    & \multicolumn{1}{c}{17.84} 
    & \multicolumn{1}{c}{\textbf{13.64} }
    & \multicolumn{1}{c|}{\textbf{11.42}} %
    & \multicolumn{1}{c}{4.73} 
    & \multicolumn{1}{c}{2.69} 
    & \multicolumn{1}{c|}{2.50}
    & \multicolumn{1}{c}{5.31} 
    & \multicolumn{1}{c}{2.82} 
    & \multicolumn{1}{c}{2.74}\\
    
\multicolumn{2}{l|}{+MonoSKD}
    & \multicolumn{1}{c}{\textbf{27.53}} 
    & \multicolumn{1}{c}{\textbf{18.25}}
    & \multicolumn{1}{c|}{\textbf{14.96}}
    & \multicolumn{1}{c}{\textbf{36.15}} 
    & \multicolumn{1}{c}{\textbf{25.08}} 
    & \multicolumn{1}{c|}{\textbf{21.14}} %
    & \multicolumn{1}{c}{\textbf{16.24} }
    & \multicolumn{1}{c}{\textbf{11.92} }
    & \multicolumn{1}{c|}{9.28}
    & \multicolumn{1}{c}{\textbf{18.43}} 
    & \multicolumn{1}{c}{13.60} 
    & \multicolumn{1}{c|}{10.77} %
    & \multicolumn{1}{c}{\textbf{5.50} }
    & \multicolumn{1}{c}{\textbf{3.45} }
    & \multicolumn{1}{c|}{\textbf{3.00}}
    & \multicolumn{1}{c}{\textbf{6.48} }
    & \multicolumn{1}{c}{\textbf{3.76} }
    & \multicolumn{1}{c}{\textbf{3.59}}\\

\hline
\end{tabular}
\end{table*}

\subsection{Detailed Car/Pedestrian/Cyclist Detection}
Compared to the 'Car' category, the 'Pedestrian' and 'Cyclist' categories have small sizes, non-rigid structures, and limited training samples, so they are much more challenging to detect.
As shown in Table \ref{Appendix_ped_cyc_car}, we report the full detection and BEV performance for Pedestrian, Cyclist and Car categories, and we can see that our scheme is still effective.
To avoid ambiguity, we clarify that DID-M3D is only trained in the Car category in the main text, so Car's performance will be slightly higher. 
In order to compare with the pre-trained model provided by DID-M3D, we retrain the models of three categories.

Observing Table \ref{Appendix_ped_cyc_car}, our MonoSKD is better on almost all metrics but slightly weaker on 'Hard' metrics for the pedestrian category. 
It is reasonable because we scale the feature map during distillation to speed up the convergence of Spearman loss, so the pedestrian category does not get enough training due to its smaller size. 
In addition, we find that our scheme has apparent advantages at almost all 'Easy' and 'Mod.' levels, proving our point of view from the side.
We will talk about this in the next section.

\subsection{Weaknesses of Our Method}
Although our scheme has achieved performance improvements on the KITTI \textit{validation} set and \textit{testing} set, it cannot be ignored that our scheme still has some shortcomings.

We adapt the scaling strategy because the detection task is a dense prediction task, so too many feature pixels lead to a decrease in sorting efficiency. 
In order to speed up the training, we adapt the scaling strategy.
Therefore, small objects that originally accounted for a low proportion of the feature map may disappear entirely after scaling.
Consequently, our method does not all outperform baselines, especially in 'Hard' settings.

\textbf{Suggestions.}~
Limited by the currently almost unsolvable sorting efficiency problem, we recommend that when using Spearman distillation, determine the feature map size in the distillation process according to the data set or only distill the object area in the feature map.
In our experiments, we find that by increasing the size of the feature map during Spearman distillation or focusing on the features of the object region without scaling, performance improvement can be obtained.
For the former, if the categories are large objects (such as "car" in KITTI), relatively small feature map sizes can be used to speed up training. 
Otherwise, training time will be sacrificed.
For the latter, we only need to use the features of the region of interest without scaling.
Considering efficiency issues, our experimental results in the main text do not adopt these suggestions.

\begin{table}
\centering
\caption{
\textbf{Comparing Pearson and Spearman correlation coefficients on the KITTI \textit{validation} set.}
We choose \textbf{MonoDLE} as baseline model.
'PCC' stands for Pearson Correlation Coefficient, and 'SCC' stands for Spearman Correlation Coefficient.
The best results are listed in bold.
}
\label{table_pcc_scc}
\begin{tabular}{c|c|c|c|c|c|c|c}
\hline
    \multicolumn{2}{c|}{\multirow{2}{*}{Methods}}
    &\multicolumn{3}{c|}{$AP_{3D}$(Car val)}
    &\multicolumn{3}{c}{$AP_{BEV}$(Car val)}\\

\multicolumn{2}{c|}{}
    & \multicolumn{1}{c}{Easy} 
    & \multicolumn{1}{c}{Mod.} 
    & \multicolumn{1}{c|}{Hard}
    & \multicolumn{1}{c}{Easy} 
    & \multicolumn{1}{c}{Mod.} 
    & \multicolumn{1}{c}{Hard}\\
\hline
\hline
\multicolumn{2}{l|}{MonoDistill}
    & \multicolumn{1}{c}{24.40} 
    & \multicolumn{1}{c}{18.47} 
    & \multicolumn{1}{c}{16.46}
    & \multicolumn{1}{|c}{32.86} 
    & \multicolumn{1}{c}{25.14} 
    & \multicolumn{1}{c}{21.99}\\
\multicolumn{2}{l|}{with PCC}
    & \multicolumn{1}{c}{25.08} 
    & \multicolumn{1}{c}{18.69} 
    & \multicolumn{1}{c|}{16.58}
    & \multicolumn{1}{c}{33.66} 
    & \multicolumn{1}{c}{25.30} 
    & \multicolumn{1}{c}{22.11}\\
\multicolumn{2}{l|}{with SCC}
    & \multicolumn{1}{c}{\textbf{26.10}}
    & \multicolumn{1}{c}{\textbf{19.18}}
    & \multicolumn{1}{c|}{\textbf{16.96}}
    & \multicolumn{1}{c}{\textbf{34.77}}
    & \multicolumn{1}{c}{\textbf{25.75}}
    & \multicolumn{1}{c}{\textbf{22.44}}\\
\hline
\end{tabular}
\end{table}

\subsection{Discussion about Pearson Knowledge Distillation}\label{Discussion about PKD}
Our scheme uses the Spearman correlation coefficient. 
Someone may wonder why we do not select the Pearson correlation coefficient for distillation.
This section focuses on applying the Pearson correlation coefficient in cross-modal distillation.

First of all, PKD-based distillation strategies are not suitable for cross-modal tasks.
Although the normalization operation can further alleviate the feature difference, this method is not the optimal solution under the premise of a vast modal difference.
In the 2D object detection task, due to the input data's consistent modality, the whole task's distillation is relatively easy, even for heterogeneous 2D detectors.

Besides, PKD is a particular distillation method that considers both relation-based and feature-based distillation.
We think PKD is more inclined to feature-based distillation and replace the feature map L1 loss in MonoDistill with PKD.
As is shown in Table \ref{table_pcc_scc}, the PCC strategy can achieve a certain performance improvement in the \textbf{\textit{validation}} set, but the improvement is very limited.
In contrast, our scheme brings a more obvious performance improvement.
In Table \ref{table_PKD_test}, we show the performance results of our scheme and PKD on the KITTI \textbf{\textit{testing}} set. 
The results show that our method significantly outperforms PKD.

\begin{table}
\setlength\tabcolsep{4.5pt}
\centering
\caption{
    \textbf{Comparing Pearson and Spearman correlation coefficients on the KITTI \textit{testing} set.}
    We choose \textbf{MonoDLE} and \textbf{DID-M3D} as baseline model.
    The best results are listed in bold.
}
\label{table_PKD_test}
\begin{tabular}{c|c|c|c|c|c|c|c}
\hline
    \multicolumn{2}{c|}{\multirow{2}{*}{Methods}}
    &\multicolumn{3}{c|}{$AP_{3D}$(Car test)}
    &\multicolumn{3}{c}{$AP_{BEV}$(Car test)}\\
\multicolumn{2}{c|}{}
    & \multicolumn{1}{c}{Easy} 
    & \multicolumn{1}{c}{Mod.} 
    & \multicolumn{1}{c|}{Hard}
    & \multicolumn{1}{c}{Easy} 
    & \multicolumn{1}{c}{Mod.} 
    & \multicolumn{1}{c}{Hard}\\
\hline
\hline
\multicolumn{2}{l|}{PKD+MonoDLE}
    & \multicolumn{1}{c}{23.39} 
    & \multicolumn{1}{c}{16.51} 
    & \multicolumn{1}{c}{14.08}
    & \multicolumn{1}{|c}{31.92} 
    & \multicolumn{1}{c}{22.03} 
    & \multicolumn{1}{c}{19.90}\\
\multicolumn{2}{l|}{MonoSKD+MonoDLE}
    & \multicolumn{1}{c}{\textbf{24.75}}
    & \multicolumn{1}{c}{\textbf{17.07}}
    & \multicolumn{1}{c|}{\textbf{14.41}}
    & \multicolumn{1}{c}{\textbf{34.43}}
    & \multicolumn{1}{c}{\textbf{23.62}}
    & \multicolumn{1}{c}{\textbf{20.59}} \\
\midrule
\multicolumn{2}{l|}{PKD+DID-M3D}
    & \multicolumn{1}{c}{26.55} 
    & \multicolumn{1}{c}{16.89} 
    & \multicolumn{1}{c}{14.74}
    & \multicolumn{1}{|c}{36.01} 
    & \multicolumn{1}{c}{23.71} 
    & \multicolumn{1}{c}{20.25}\\
\multicolumn{2}{l|}{MonoSKD+DID-M3D}
    & \multicolumn{1}{c}{\textbf{28.43}}
    & \multicolumn{1}{c}{\textbf{17.35}}
    & \multicolumn{1}{c|}{\textbf{15.01}}
    & \multicolumn{1}{c}{\textbf{37.12}}
    & \multicolumn{1}{c}{\textbf{24.08}}
    & \multicolumn{1}{c}{\textbf{20.37}}  \\
\hline
\end{tabular}
\end{table}

\subsection{More Qualitative Results}\label{More Qualitative Results}

\begin{figure*}[t]
\centering {\includegraphics[width=0.323\textwidth]{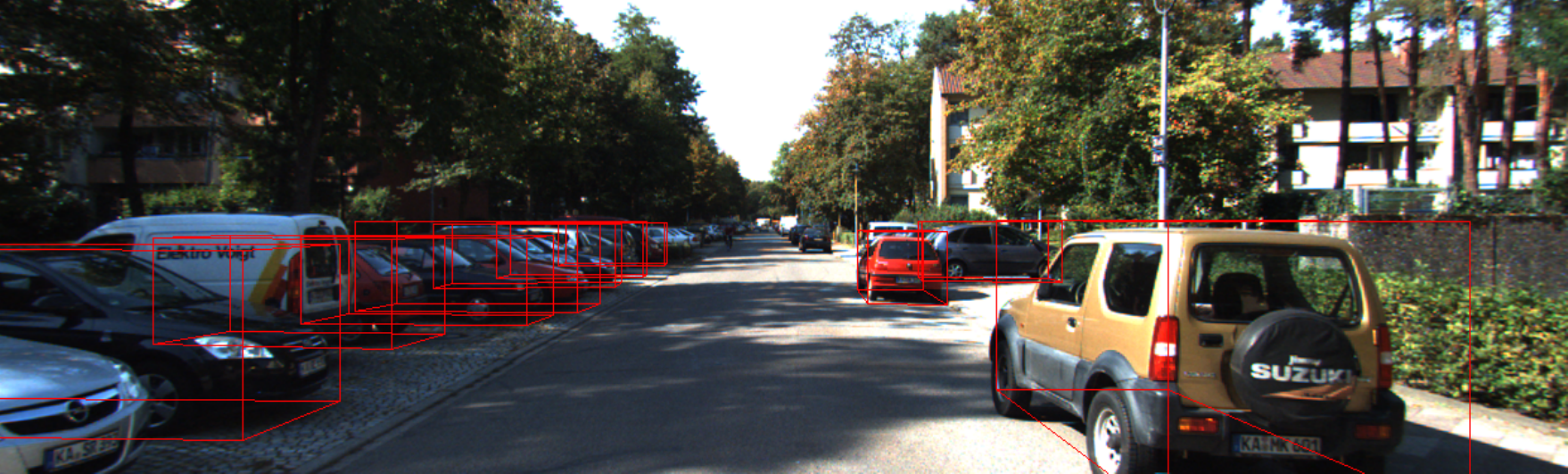}}
\centering {\includegraphics[width=0.323\textwidth]{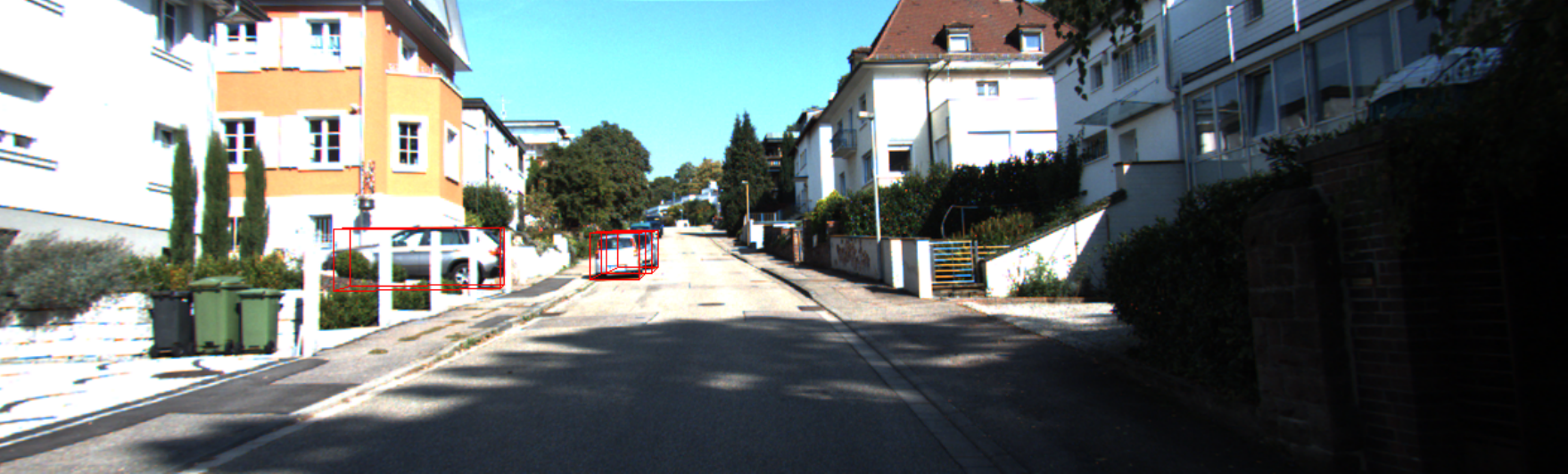}}
\centering {\includegraphics[width=0.323\textwidth]{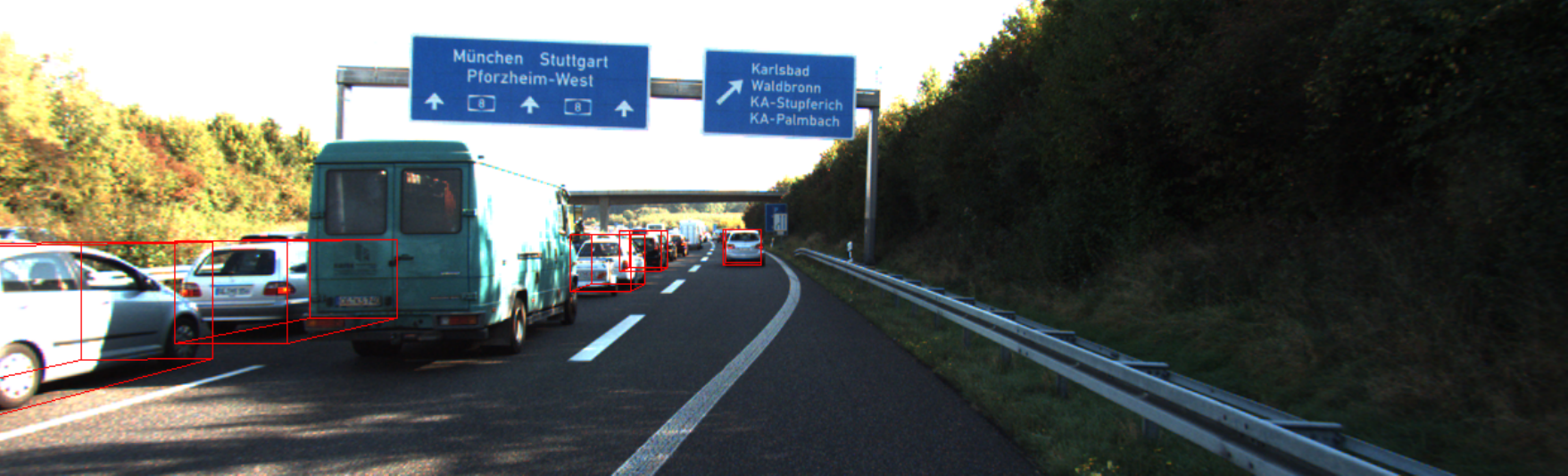}}
\\
\centering {\includegraphics[width=0.323\textwidth]{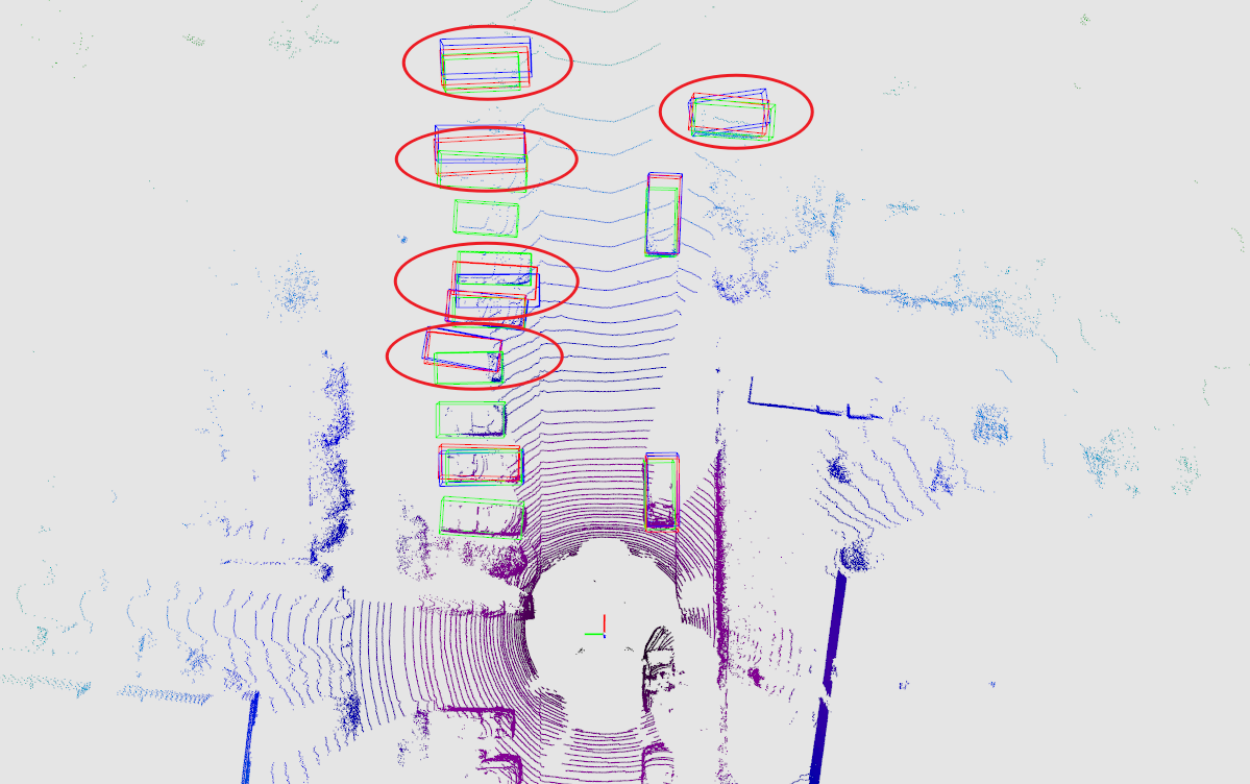}}
\centering {\includegraphics[width=0.323\textwidth]{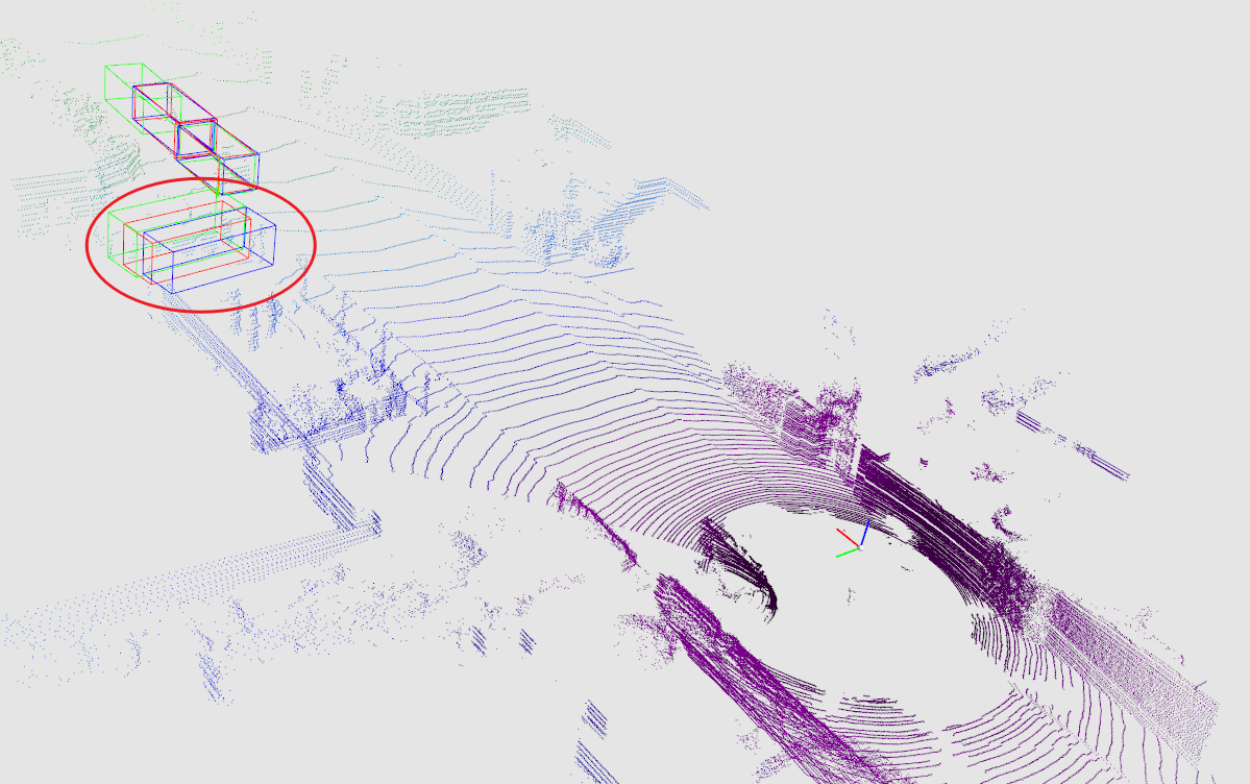}}
\centering {\includegraphics[width=0.323\textwidth]{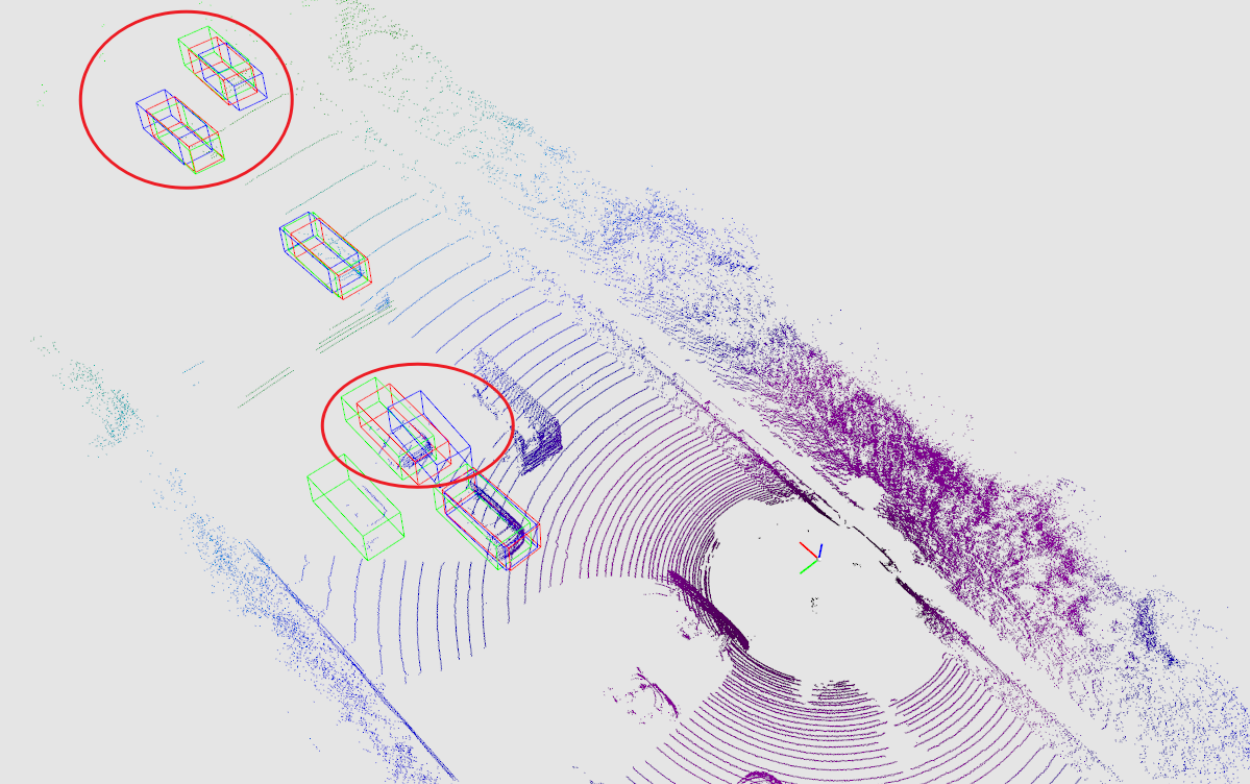}}
\\
\vspace{5pt}
\centering {\includegraphics[width=0.323\textwidth]{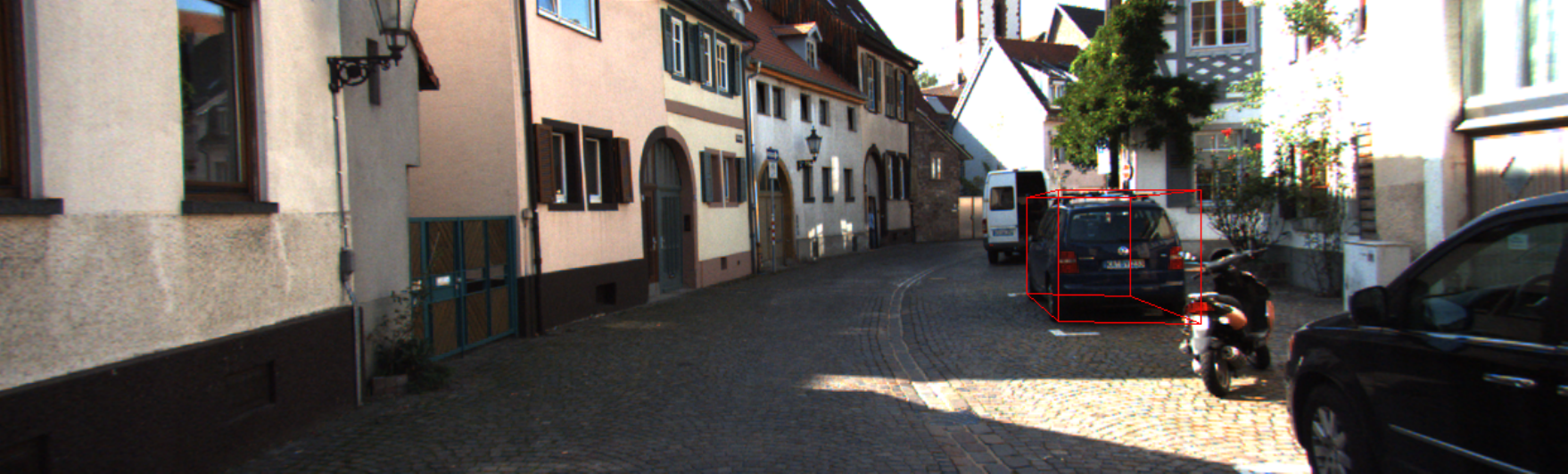}}
\centering {\includegraphics[width=0.323\textwidth]{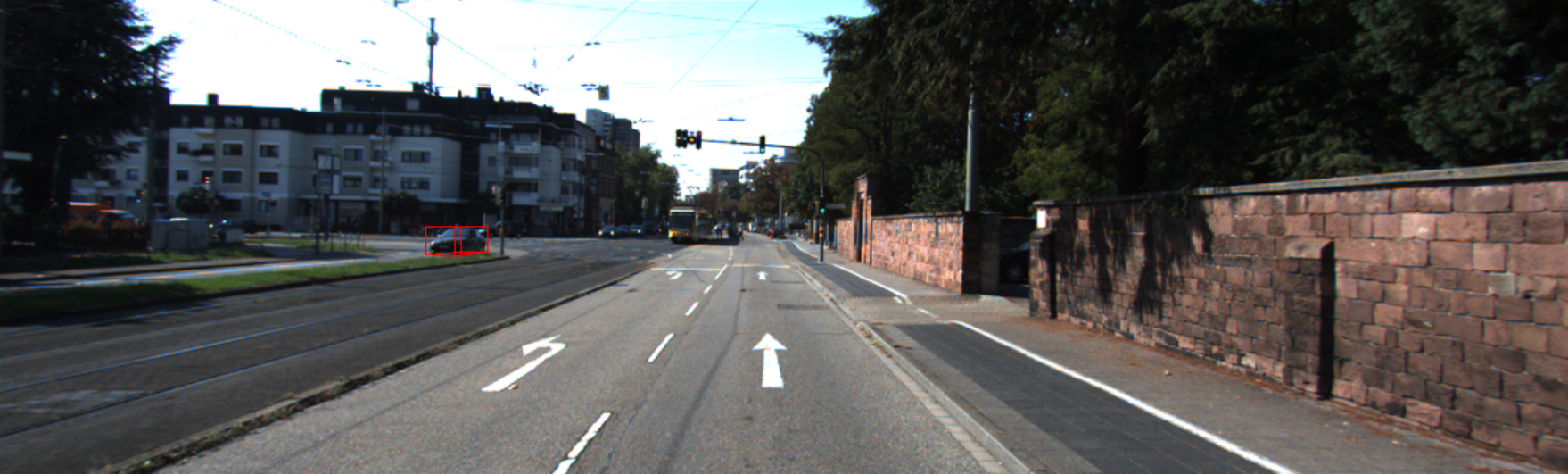}}
\centering {\includegraphics[width=0.323\textwidth]{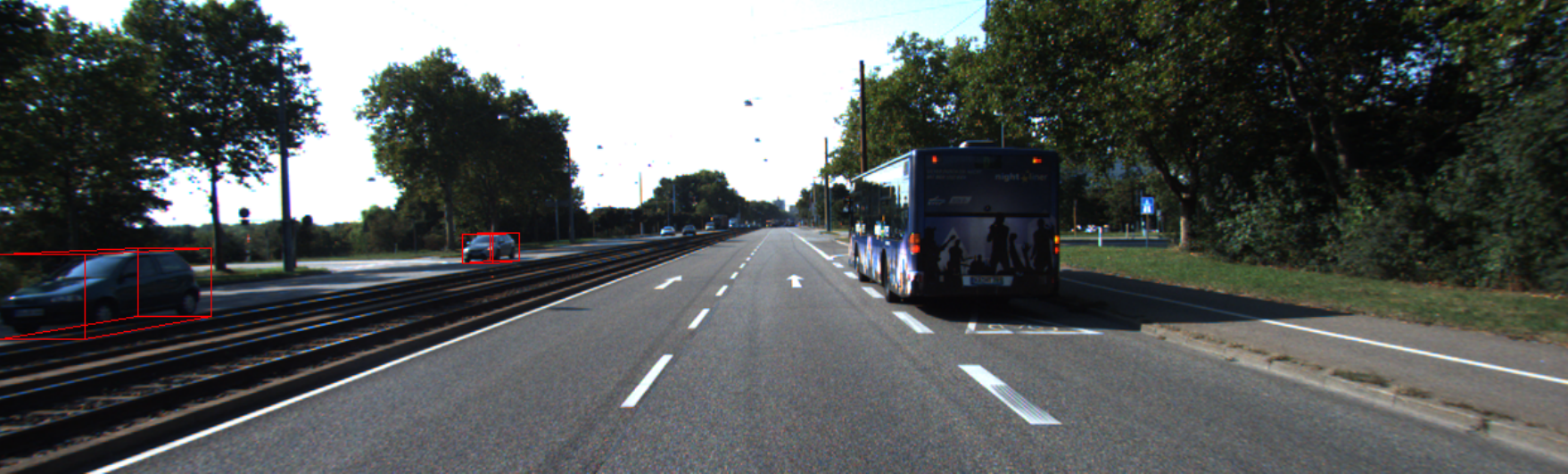}}
\\
\centering {\includegraphics[width=0.323\textwidth]{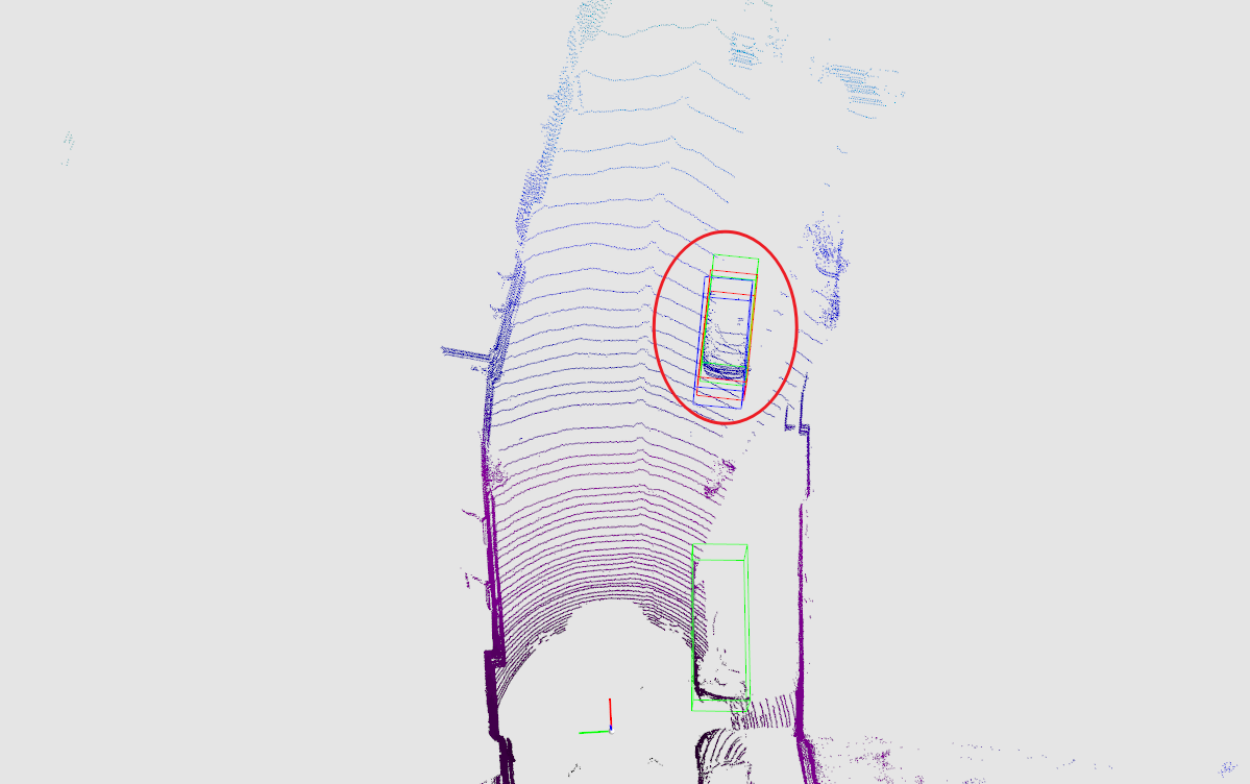}}
\centering {\includegraphics[width=0.323\textwidth]{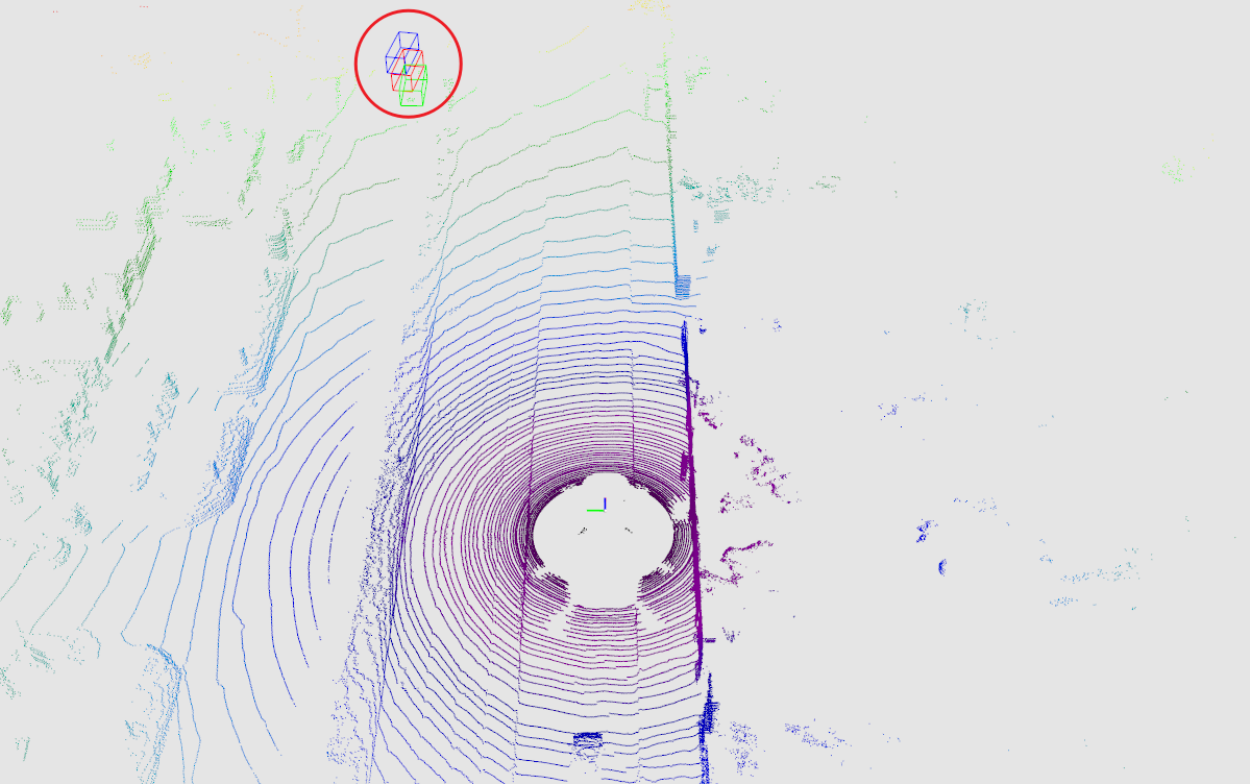}}
\centering {\includegraphics[width=0.323\textwidth]{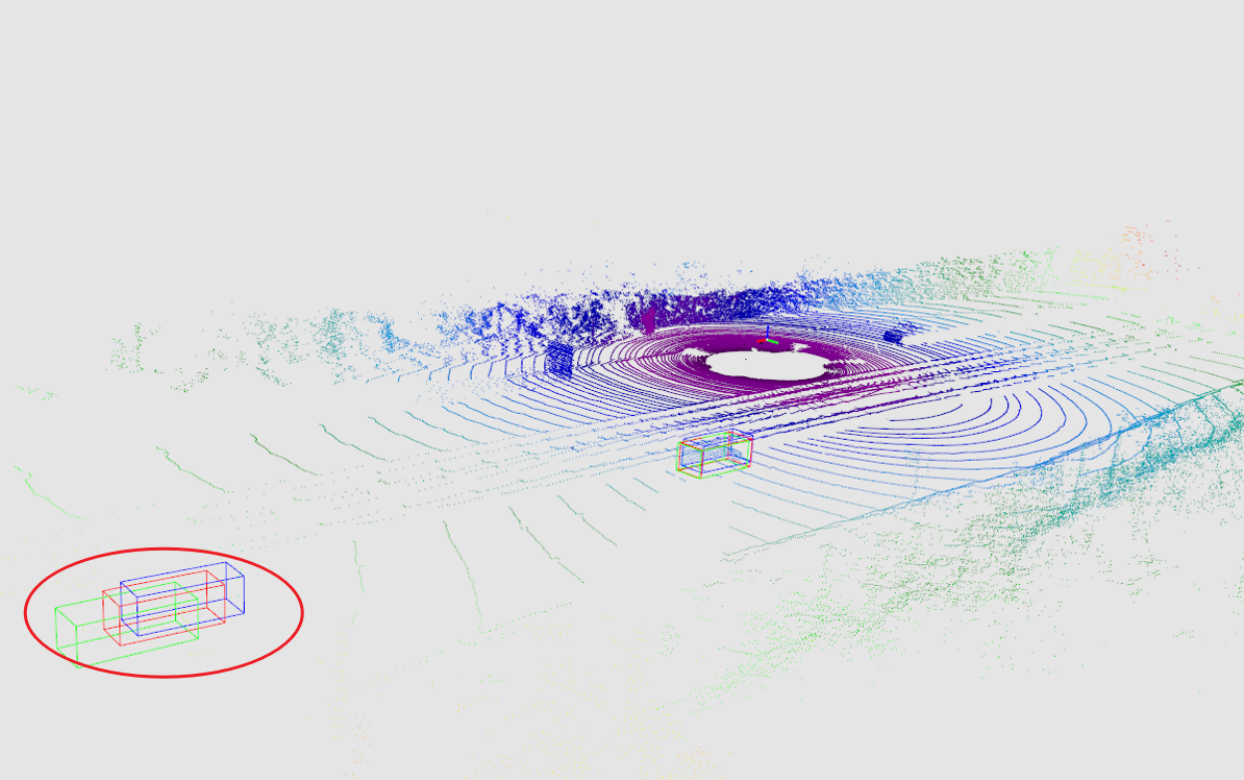}}
\caption{
\textbf{More qualitative results.}
We use blue, red, and green 3D boxes to denote the DID-M3D baseline, MonoSKD, and ground truth results. 
Additionally, we use red circles to highlight significant differences.
}
\label{fig5}
\end{figure*}

Due to the space limitation of the text, we have added more quantitative results in the appendix.
As is shown in Figure \ref{fig5}, our prediction results are closer to the gt box than the baseline, so we have a more powerful performance.

\end{document}